# AI Consciousness is Inevitable:

## A Theoretical Computer Science Perspective

Lenore Blum and Manuel Blum

ABSTRACT

We look at consciousness through the lens of Theoretical Computer Science (TCS), a branch of mathematics that studies computation under resource limitations, distinguishing functions that are efficiently computable from those that are not. From this perspective, we are developing a *formal* TCS *model* for consciousness. We are *inspired* by Alan Turing's simple yet powerful model of computation and Bernard Baars' theater model of consciousness. Though extremely simple, the model (1) aligns at a high level with many of the major scientific theories of human and animal consciousness, (2) provides explanations at a high level for many phenomena associated with consciousness, (3) gives insight into how a machine can have subjective consciousness, and (4) is clearly buildable. This combination supports our claim that machine consciousness is not only plausible but inevitable.

Lenore Blum
lblum@cs.cmu.edu

Manuel Blum
mblum@cs.cmu.edu

# 1 Introduction

We study consciousness from the perspective of **Theoretical Computer Science (TCS)**, a branch of mathematics concerned with understanding the underlying principles of computation and complexity, including especially the implications and surprising consequences of resource limitations. As a branch of mathematics, the TCS perspective abstracts and formalizes informal concepts, proposes axioms and derives results from them.[1]

From this perspective, we are *developing* a *formal* TCS *model* for consciousness. Our aim is to start with a reasonable and minimal set of formal assumptions about consciousness, then see what follows: the pros and cons of the model. We have tried to make our model "Einstein's razor" simple.

The **Conscious Turing Machine (CTM)** is a *simple formal machine* model of *consciousness* inspired in part by Alan Turing's *simple,* yet powerful, *formal machine* model of *computation* (Turing, 1937)**,** and by Bernard Baars' *theater model* of *consciousness*, also called the *global workspace* (**GW**) theory of consciousness (Baars, 1997).[2] In the spirit of TCS, informal notions are defined formally *in the* CTM.

In (Blum & Blum, 2022), we consider how a CTM could exhibit various phenomena associated with consciousness (e.g., blindsight, inattentive blindness, change blindness) and present CTM explanations that agree, at a high level, with cognitive neuroscience literature.

---

[1] Theoretical Computer Science, a field started in the 1960's, grew out of the Theory of Computation (TOC), a field with origins in the 1930's. By taking resource limitations into account, the TCS perspective differentiated itself from TOC where limitations of time and space do not figure. TOC distinguishes computable from not computable. It does not distinguish between efficiently (feasibly) computable and not efficiently (not feasibly) computable. For a brief history of TOC and TCS, see Appendix, section 6.1.

[2] The *global workspace* has origins in architectural models of cognition, developed largely at Carnegie Mellon University by: Herb Simon's Sciences of the Artificial (Simon, 1969), Raj Reddy's Blackboard Model (Reddy, 1976), Allen Newell's *Unified Theories of Cognition* (Newell, 1990) and John Anderson's ACT-R (Anderson, 1996). Indeed, (Baars, 1997) states that: "Global Workspace theory derives from the integrative modeling tradition of Alan Newell, Herbert Simon, John Anderson, and others in cognitive science."

In contrast to Turing, we take resource limitations into account, both in designing the CTM model (e.g., size of chunks, speed of computation) and in how resource limitations affect (and help explain) phenomena related to consciousness (e.g., focus and dreams) and related topics such as the paradox of free will (Blum & Blum, 2022).

Our perspective differs even more. What gives CTM its feeling of consciousness is not its input-output map, nor its computing power, but what's under the hood.[3] In this paper we take a brief look under the hood (section 2.1).

David Chalmers' introduction of the Hard Problem (Chalmers, 1995) helped classify most notions of consciousness into one of two types. The first type, variously called *access* (Block, 1995) or *functional* or *cognitive* consciousness (Humphrey, 2023), we call *conscious attention* (Formal Definition 1 in section 2.2).[4] In CTM, conscious attention is due to the simultaneous reception by all processors of the global broadcast.

The second type (associated with the Hard Problem) is called *subjective* or *phenomenal* consciousness (Nagel T. , 1974) and is generally associated with sensory experiences or *qualia*, (Peirce, 1866), modified by (Lewis, 1929). We call it *conscious awareness* (Formal Definition 9 in section 2.3.2.2). Conscious awareness in CTM is a consequence of the co-evolution of the *Model-of-the-World* with *Brainish*, its self-generated multimodal/multisensory internal language.

We view Chalmers' Hard Problem as a challenge to show that the latter, subjective consciousness, is "computational".[5]

---

[3] This is important. We claim that simulations that modify CTM's key internal structures and processes will not in general experience what CTM experiences. On the other hand, we are not claiming that the CTM is the only possible machine model to experience *feelings of consciousness*. The CTM is *a minimal* formal machine model for consciousness. Indeed, Wanja Wiese describes the CTM as a "minimal unifying model", (Wiese, 2020) and personal communication.

[4] In previous papers, e.g., (Blum & Blum, 2022), we used the words "conscious awareness" to denote what we formally define in this paper as *conscious attention*.

[5] We are not alone in this view, see e.g., (Dehaene, Charles, King, & Marti, 2014).

Specifically, we contend that a machine that interacts with its worlds (inner and outer) via input sensors and output actuators, that constructs models of these worlds enabling planning, prediction, testing, and learning from feedback, and that develops a rich internal multimodal/ multisensory language, can have both types of consciousness. Again, we contend that subjective consciousness is computational.

We emphasize that the CTM *is a simple formal machine model* designed to explore and understand consciousness from a TCS perspective. While it is inspired by and owes much to cognitive and neuroscience theories of consciousness, CTM *is not intended* to model the brain *nor* its neural correlates of consciousness. It is way too simple for that.

And, although the CTM is inspired by cognitive neuroscientist Bernard Baars' global workspace (GW) theory of consciousness (Baars, Bernard J., 1997), it ***is not*** a standard GW model. The CTM *differs* from GW in a number of important ways:

- In CTM, *competition* for global broadcast is *formally defined* (sections 2.1.3 and 6.2).
- The CTM does away completely with the ill-defined Central Executive of other GW models.
- In CTM, a distributed *Model-of-the-World processor* (section 2.1.2.1) constructs and employs (distributed) *models* of its (*inner* and *outer*) worlds (section 2.3.1) *to make sense* of itself and itself in its worlds.
- *Brainish* (sections 2.1.2.1 and 2.3.1), CTM's *self-generated multimodal internal language*, enables communication between processors and provides the multimodal labeling of sketches in CTM's world models. The co-evolution of Brainish and world models play important roles in generating CTM's subjective feelings such as pain and pleasure (sections 2.3.1.1and 2.3.1.2).
- In CTM, *predictive dynamics* (cycles of prediction, testing, feedback and learning, locally and globally) constantly improves its world models (see section 2.1.8).

The CTM interacts with the world via input *sensors* and output *actuators* (section 2.1.6), enabling the creation of its *embodied* (Shanahan, 2010), *embedded, enacted* (Maturana & Varela, 1972) and *extended mind*[6] (Clark & Chalmers, 1998).

---

[6] In principle, we can also allow off-the-shelf processors in our CTM.

CTM naturally *aligns* with and *integrates* features considered key to human and animal consciousness by many of the major scientific theories of consciousness (Section 3).[7] These theories consider different aspects of consciousness and often compete with each other (Lenharo, 2024). Yet the alignment of their underlying principles with CTM at a high level helps demonstrate their compatibility[8].

Even more, their *alignment* with CTM, *a simple machine model* that exhibits many of the important phenomena associated with consciousness, supports our claim that a conscious AI is inevitable.[9, 10]

While working on this paper, we became aware of Kevin Mitchell's blog post in *Wiring the Brain* (Mitchell, 2023) in which he makes a point similar to one that we make, namely, that many of the major theories of consciousness are compatible and/or complementary[11]. For a similar conclusion, see (He, 2023) and (Storm, et al., 2024). Even more, Mitchell presents fifteen questions "that a theory of consciousness should be able to encompass". He declares that "even if such a theory can't currently answer all those questions, it should at least provide an *overarching framework* [12] (i.e., what a theory really should be) in which they can be asked in a coherent way, without one question destabilizing what we think we know about the answer to another one."

---

[7] These theories include: The Global Workspace/Global Neuronal Workspace (GW/GNW), Attention Schema Theory (AST), the Self-Organizing Metarepresentational Account (SOMA), Predictive Processing (PP), Integrated Information Theory (IIT), Embodied, Embedded, Enacted and Extended (EEEE) theories, Evolutionary theories, and the ERTAS (Extended Reticulothalamic Activating System) + FEP (Free Energy Principle) theories.

[8] We are often asked: how can these competing theories be compatible? By taking the TCS approach and abstracting these theories' underlying principles, we are not mapping theories to different parts of the human brain as is done in the adversarial competitions (Lenharo, 2024) and (Cogitate, Ferrante, Gorska-Klimowska, & al., 2025).

[9] Thus, our response to the query, "could machines have it [consciousness]?" (Dehaene, Lau, & Kouider, 2017), is "YES it could". (Farisco, Evers, & Changeux, 2024) consider the question "Is artificial consciousness achievable?" from the perspective of the human brain. They conclude with a tentative possibility for "non human-like forms of consciousness."

[10] Additionally, supporting the view that consciousness is intimately tied to intelligence, we indicate how CTM provides a *framework* for constructing an Artificial General Intelligence (AGI). (See section 2.4.)

[11] We use "complementary" to emphasize the idea that the diverse theories can be seen as different pieces of a larger puzzle rather than conflicting viewpoints.

[12] Italics are ours.

Mitchell's questions are thoughtful, interesting, and important. Later in this paper (Appendix, section 6.3), we offer preliminary answers from the perspective of CTM. Our answers to Mitchell's questions supplement and highlight material in the following overview of CTM.[13]

Finally, the development of CTM is a work in progress. Indeed, we view this paper as an extended outline for our forthcoming monograph (Blum, Blum, & Blum, 2026).

# 2 The Conscious Turing Machine (CTM)

## 2.1 Structure (STM, LTM, Up-Tree, Down-Tree, Links, Input, Output) and Predictive Dynamics

**CTM** is defined formally as a 7-tuple, (STM, LTM, Up-Tree, Down-Tree, Links, Input, Output). These seven components each have well-defined properties (Blum & Blum, 2022) which we outline in this section.

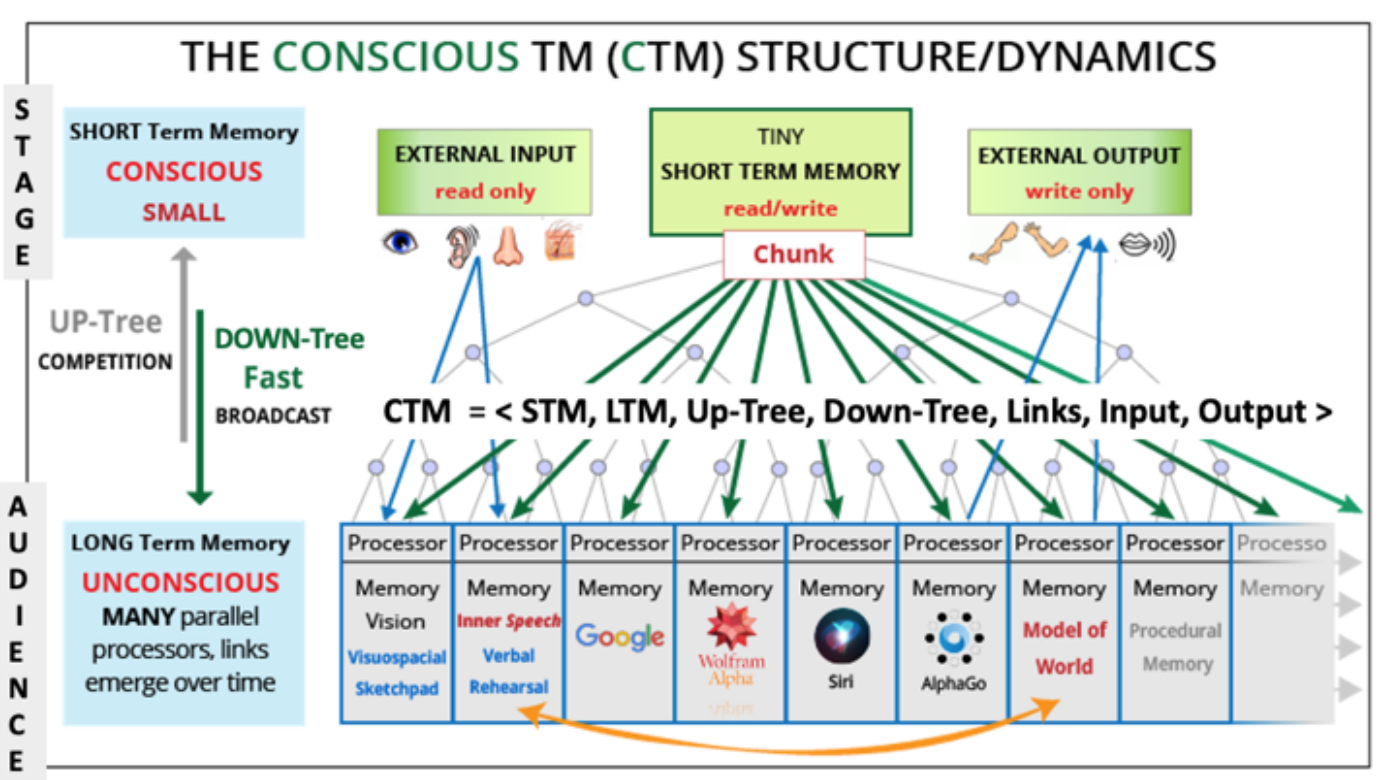

**Figure 1. Sketch of the Conscious Turing Machine**

[13] In the overview (Section 2), we annotate paragraphs that refer to Kevin Mitchell's queries. For example, if a paragraph has a label [KM1], then it refers to Mitchell's first question, KM1.

To start, CTM has a *finite* lifetime **T.** Time **t = 0, 1, 2, 3, … , T** is measured in discrete clock ticks. **T** is a parameter.

CTM has both a time-varying ***inner world*** **I(t)** and a time-varying ***outer world*** **O(t).** Formally, these are state spaces. Informally, CTM's *inner world* includes its internal mechanisms and processes, and its "thoughts" and memories; its *outer world* is where it lives, what's outside CTM.

Again, to emphasize that CTM is *embodied*, *embedded, enactive* and *extended* in its worlds (inner and outer), we should denote it by **rCTM** (**r** for robot).[14] Instead, we use the acronym CTM for both CTM and rCTM. We now outline the properties of its components:

### 2.1.1 STM, a Buffer and Broadcasting Station

The stage in the theater model is represented in CTM by what we call **Short Term Memory (STM)** (though it is not much of a memory at all). STM is not a processor; it is merely a buffer and broadcasting station. STM can hold only *one* "chunk of information" at a time. (For the definition of *chunk*, see section 2.1.3.) One is not the "magical number" $7 \pm 2$ (Miller G. A., 1956), but we are looking for simplicity and a single chunk will do.[15]

### 2.1.2 LTM Processors

The audience, which we call **Long Term Memory (LTM)**, is represented by a massive collection of **N** (initially independent) *powerful* processors that comprise CTM's *principal computational machinery* and *long term memory*. **N** is a parameter: for our particular CTM, we choose $\mathbf{N = 2^{24}}$.[16]

---

[14] As AI's continue to interact with the world, they also (increasing) incorporate the 4 E's.

[15] The "magical number" $\mathbf{7 \pm 2}$ was proposed by George Miller (Miller G. A., 1956) as the number of "chunks" that a human at any moment of time. can hold in their "short term memory". Nowadays, following (Baddeley & Hitch, 1974), that "short term memory" is called the phonological loop. In CTM, both the phonological loop and the visuo-spatial sketchpad could be LTM processors.

[16] We have chosen the numbers in our particular CTM instance so that they are (roughly) consistent with the numbers in a model of the human brain. The number **N** of LTM processors is a parameter. Choosing $\mathbf{N = 2^{24}}$ for our particular instance is suggested by the $\mathbf{10^7}$ ($\mathbf{\sim 2^{23.25}}$) cortical columns in the human brain. Two times that number accounts for additional important processors. (In addition, $\mathbf{2^{24}}$ rather than $\mathbf{2^{23}}$ makes it easier for us humans to do computations in your head than $\mathbf{2^{23}}$.)

These processors are *random access machines* (*not* Turing machines).[17] We assume that each processor has its own internal time clock for running its own algorithms, and that that time clock is **100** times faster than CTM's global time clock. (For more on processor properties, see section 2.1.7.)

In CTM, all processors are in LTM so when we speak of a processor, we mean an LTM processor.[18] These processors are "unconscious".

All processors create *chunks* of information (section 2.1.3) to communicate within the CTM. They compete to get their chunks on stage to be immediately broadcast to an attentive audience.[19]

---

CTM's lifetime **T** is a parameter. If we choose **CTM**'s lifetime $\mathbf{T = 2^{10}N = 2^{34}}$ $\mathbf{(\sim 10^3 N \sim 10^{10})}$ ticks, and a clock frequency of $\mathbf{2^3}$ ticks per second (~ the alpha frequency of the brain), **CTM**'s lifetime is about **68** years. We get the same lifetime of **68** years if $\mathbf{T = 2^{13}N}$ ticks and clock frequency is $\mathbf{2^6}$ ticks per second (~ the gamma frequency of the brain).

[17] A random access machines (RAM) can search a sorted list in logarithmic time; a Turing Machine needs linear time.

CTM's finite lifetime imposes a finite lifetime on LTM processors. However, for understanding consciousness, processors are better modeled as RAMs *running algorithms* rather than as finite state machines executing tuples.

[18] Hence, the processors that Baars puts outside of LTM, like the phonological loop, visuospatial sketchpad (Baddeley & Hitch, 1974), and episodic buffer (Baddeley, 2000) will be in CTM's LTM.

[19] In Baars' theater metaphor, consciousness is the activity of actors in a play performing on a stage of Working or Short Term Memory (STM). The Inner Speech actor is often on stage. Their performance is under observation by a huge audience of powerful unconscious processors in Long Term Memory (LTM) that are sitting in the dark. The unconscious processors vie amongst themselves to get their script/query/information on stage to be broadcast to the audience.

As an example of the theater metaphor, consider the **"What's her name?"** scenario:

Suppose at a party, we see someone we know but cannot recall her name. Greatly embarrassed, we rack our brain to remember. An hour later when driving home, her name pops into our head (unfortunately too late). What's going on?

Suppose we have a CTM brain. Racking our brain to remember caused the *urgent* request **"What's her name?"** coming from LTM processor **p** to rise to the stage (STM), which immediately broadcasts the question to the audience. Many (LTM) processors try to answer the query. One such processor recalls we met in a neuroscience class; this information gets to the stage and is broadcast, triggering another processor to recall that what's-her-name is interested in "consciousness", which is broadcast. Another processor **p'** sends information to the stage asserting that her name likely begins with **S.**

#### 2.1.2.1 The Model-of-the-World Processor (MotWp), The Model-of-the-World (MotW), and Brainish

We single out what we call the ***Model-of-the-World processor*** (**MotWp**), which is not actually a single processor; its functionality and memory are distributed across all LTM processors. It constructs models of CTM's inner and outer worlds, which it stores in these processors. We call this collection of constructed models the ***Model-of-the-World*** **(MotW)**. The MotW starts off as a "foggy blob"; over time it becomes more sharply defined and populated with many labeled sketches.

While each processor has its own distinct language, the (community of LTM) processors communicate with each other in ***Brainish***, CTM's internal self-generated multimodal inner language. A ***Brainish gist***[20] is a "succinct phrase" consisting of a finite sequence of at most **8** Brainish words,[21] each word being a pair **<p, t>**, where **p** is a processor number and **t** is a time in clock ticks. A Brainish gist ***fuses*** modalities (e.g., sight, sounds, smells, sense of touch) and processes (e.g., thoughts like, "to prove this math statement, use induction"). (See section 2.3 and its subsections, in particular section for formal definitions and development of these notions.)

---

Sometime later the stage receives information from processor **p’’** that her name more likely begins with **T**, which prompts processor **p’’’** (which like all processors has been paying attention to all the broadcasted information) to claim with *great certainty* (and correctly) that her name is **Tina**. The name is broadcast from the stage, our audience of processors receives it, and we finally consciously remember her name. Our conscious self has no idea how or where her name was found.

[20] We were unaware of Jeremy Wolfe's 1998 paper (Wolfe, 1998) until we read Fei-Fei Li's wonderful book *The Worlds I See* (Li, 2023), where she identifies Wolfe as the "gist" guy. Wolfe uses the word "gist" in reference to "what we remember – and what we forget – when we recall a scene." Our explanation of *change blindness* in CTM (Blum & Blum, 2022) is almost identical to Wolfe's explanation of change blindness in humans: the same gist describes both the original and changed scenes.

[21] We could require gists to be at most 2 or 3 Brainish words instead of **8**, but for creating non-trivial examples, **8** is easier (and the maximum for quick computations, as in the Up-Tree competition).

A Brainish gist (phrase) is like a dream or movie frame. The Brainish language evolves over CTM's lifetime, from nothing initially into an ever-growing dictionary of words. Clearly, Brainish will differ from one CTM to another. (See section 2.3.1 on how Brainish and the MotW co-evolve.)[22]

The MotWp and its MotW play an important role in planning, predicting, testing and learning, and in CTM's *feelings of consciousness* (section 2.3). When Brainish labeled sketches are *globally broadcast* (section 2.1.4) and *unpacked* (Formal Definition 7 in section 2.3.2.2) they *evoke* CTM's unitary (**Axiom 1** in section 2.2) subjective experiences (**Axiom 2** in section 2.3.2.2).

### 2.1.3 Up-Tree Competition and Chunks

LTM processors *compete* in a well-defined fast *probabilistic competition* process to get their questions, answers, and information in the form of *chunks* onto the stage (STM). The competitions are hosted by the **Up-Tree,** a *perfect*[23] binary tree of height **h** which has a leaf in each of the $\mathbf{N = 2^h}$ LTM processors and its root in STM.

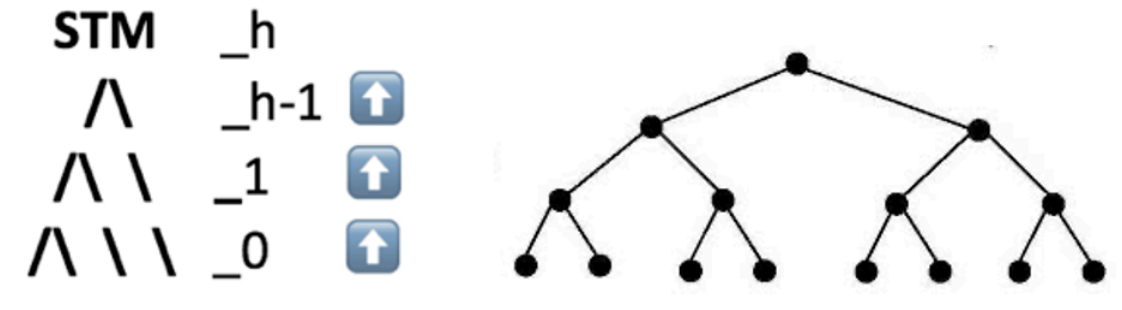

**Figure 2.** ***Imperfect*** **and** ***perfect*** **binary Up-Trees.**

At each clock tick, *a new competition starts* with each processor putting a chunk of information into its Up-Tree leaf node. Entry chunks compete with their *siblings*. (Chunks are siblings if they are at nodes that are children of the same parent.) Winning siblings move up the binary tree (losing siblings drop out). Local winners compete at the next tick with their new siblings (as in a tennis or chess tournament). The process continues until a winner reaches STM at the root node. (See Appendix 6.2 for a fuller discussion of the Up-Tree competition.)

---

[22] Paul Liang is developing a computational framework for Brainish based on multimodal machine learning (Liang, 2022). This is different than CTM's evolved formal Brainish (section 2.3.1).

[23] A *perfect* binary tree is a binary tree in which all leaf nodes are at the same depth, that depth being the height **h** of the tree. The tree has $\mathbf{N = 2^h}$ leaves, and every node except the root node has a unique sibling (neighbor). For simplicity, the CTM's competition tree is a perfect binary tree with $\mathbf{h = 24, N = 2^{24}}$.

A ***chunk*** is formally defined to be a tuple,

**<address, time, gist, value; aux >**,

consisting of: the ***address***[24] of the originating processor; the ***time*** the chunk was created; a *succinct* Brainish ***gist*** (***phrase***) of information; ***value***,[25] a ***valenced*** (+/-) ***number*** **[**to indicate the importance/ urgency/ confidence/ sentiment the originating processor assigns the gist**]**; plus, some ***auxiliary information***. See Appendix, section 6.2 for a discussion of auxiliary information.

**Remark.** We note here, that as chunks move up the Up-Tree, their first four parameters will remain the same, but their ***auxiliary information*** can change, gathering more and more global information. The auxiliary information in the winning chunk will contain "global context". For example, in the ***winner-take-all*** *probabilistic competition* discussed in the Appendix, the auxiliary information in the winning chunk will have both the sum of all competing entry chunks' **intensities** (i.e., the sum of the magnitude of the entry chunks' values) and the sum of all their **moods** (i.e., of their values). Thus, in general, the auxiliary information in a winning chunk will be different than the auxiliary information when the chunk entered the competition. *Nevertheless, we may simply denote this information in both versions of the chunk with the same notation **aux**, leaving it to the reader to mentally note the revision of meaning.*

***Gists*** in chunks may convey information, pose questions, provide answers, request that specific tasks be carried out or problems solved, and so on. Brainish ***words*** and Brainish ***gists*** (*phrases*) are formally defined in section 2.3.2. The primal Brainish gist (the empty sequence) is denoted by **NIL.**

The chunk's ***value*** is estimated by the processor's ***value assigning algorithm*** (Blum, Blum, & Blum, 2026). We think of this algorithm as assessing the importance that the processor assigns the gist. That value (that the processor assigns to a chunk) is adjusted by the processor's built-in ***Sleeping Experts*** *(learning)* ***Algorithm*** (sections 2.1.7(item 3) and 2.1.8).

---

[24] The address of each of the $\mathbf{2^{24}}$ ($\mathbf{10^7 < 2^{24} < 10^8}$) processors is a **24** binary bit, or **8** decimal digit, or **3** byte number.

[25] We have previously used "weight" for "value".

Each submitted chunk competes locally with its sibling neighbor. In a single clock tick, the local winning chunk is chosen and provided it is not at the top moves up one level of the Up-Tree, ready to compete with its neighbor. At each clock tick, there are active local competitions at every node in every level from **0** to **h-1** of the **Up-Tree**. The competition that begins at time **t** ends at time **t+h** with the ***winning chunk*** in STM.[26]

Notably, for the ***winner-take-all*** *probabilistic* competition (Appendix, section 6.2), it is proved that a chunk wins the competition with probability proportional to its **f-value.**[27] As a consequence of this theorem, the winning chunk is independent of the processor's location! Clearly, this is an important feature for a machine or brain in which no movement of processors is possible. (It is a property that would be difficult if not impossible to achieve in fair tennis or chess tournaments.)

### 2.1.4 Down-Tree Broadcast, Paying Conscious Attention, and Conscious Communication

Upon entry into STM at time **t + h**, the ***winning chunk*** is immediately ***globally broadcast*** to the audience of LTM processors, which receive it at time **t+h+1**. Broadcast is via the **Down-Tree**, a bush of height **1** with root in STM and **N** leaves, one leaf in each of the **N** LTM processors. /|\ ⬇ [KM5]

Each LTM processors stores all broadcasted chunks in memory, forming a ***global (Brainish) Dictionary***. Thus, all processors have the same global dictionary.

**Remark.** In the *winner-take-all* probabilistic competition (referred to in section 2.1.3 and discussed in the Appendix, section 6.2), all processors will receive "global context" enabling them to calculate the average *intensity* and average *mood* of all chunks that entered the competition. In addition to receiving the winner's *gist*, its *value*, and its submitting *processor's address*, this will help all processors calibrate subsequent actions.

---

[26] If the interval between successive clock ticks is **.1** seconds and the number of LTM processors is $2^{24}$, then a full binary competition takes about **2.4** seconds. Define an **octal competition tree** to be a tree with **8** children per node, in which **one** level of octal tree does **three** levels of a binary tree in a single tick, **.1** seconds, rather than **.3** seconds. Then an octal tree competition takes **.8** seconds, which better mirrors what goes on in a brain.

[27] One can think of the chunk's **f-value** as being analogous to a player's ranking in tennis or chess. It would not do to pick the winner to be the chunk with the highest **f-value** for then no other chunk, even one with a slightly less **f-value**, would have a chance. The probabilistic competition gives all competitors a fair chance.

The global broadcast of a single chunk at each tick to all LTM processors enables CTM to "focus attention" on the winning gist (chunk). One is not the "magical number" **7±2**, but we are looking for simplicity and a single chunk will do.

We say that CTM ***pays conscious attention*** to the winning chunk when it is simultaneously received by all LTM processors.[28] (This is Formal Definition 1 of ***conscious attention*** in section 2.2.)

We call communication between LTM processors that goes through STM ***conscious communication***.

### 2.1.5 Links and Unconscious Communication

Links between processors enable *conscious communication*, i.e. communication that goes through STM, to be replaced by more direct and faster ***unconscious communication*** through links.

The CTM has no links (between processors) at birth. At time **t+1**, a *bi-directional* **link** is created between the processor whose chunk got broadcast at time **t** and the one whose chunk got broadcast one tick later at time **t+1**. In other words, **links** in **CTM** get formed in a "Hebbian-like" manner (Hebb, 1949), "processors that broadcast together link together". We call these links (even if they are duplicates) **fundamental.** In **CTM**'s lifetime, exactly **T fundamental links** are created.[29]

---

[28] Alison Gopnik's contention that "babies are more conscious than we are" (Gopnik, 2007) can be understood in terms of *conscious attention*. In the infant CTM, until links are formed (section 2.1.5), all communication between processors is conscious, i.e., goes through STM, and is therefore consciously attended to. On the other hand, the MotW in the infant CTM is considerably less developed than in the adult. Hence phenomenal consciousness (what we call *conscious awareness*, section 2.3) is considerably less developed in the infant CTM than in the adult.

[29] CTM has $N = 2^{24}$ LTM processors and initially no two processors are linked. If all processors were linked, there would be $2^{48}$ links, an infeasible number of links in the brain. Worse yet, a processor with $2^{24}$ inputs might need its own personal Competition Tree to decide what to pay attention to. While one might want some processors to be linked initially, we don't know which to link, so we choose for simplicity to not link any.

Adding links enables sending and receiving more information per tick. If a Brainish chunk, i.e., a chunk from the *global dictionary*, goes on a link from processor A to B and is followed (in the same tick or next) by the exact same chunk going from B to C, then a link is created from A to C. This is a second way links can be formed between processors.

The CTM has no provision for deleting links, despite that deletions are known to be important in human learning. No provision for deleting links enables the model to be simpler, and processors can still do the equivalent of deleting links by simply ignoring them.

Thus, when CTM initially learns to ride a bike, most communication is done consciously until relevant processor links have formed. Then, for the most part, riding a bike is done unconsciously until an obstacle is encountered, forcing CTM to *pay conscious attention* if its "unconscious instincts" fail to kick in.[30] [KM5]

### 2.1.6 Input/Output

The CTM is not a brain in isolation. ***Input/ Sensory processors*** take **input** from CTM's outer world via ***sensors***[31], convert that **raw input** into Brainish gists, those gists into chunks, and then submit those chunks to the competition (sections 2.1.3 and 6.2). ***Output*** **/Motor** ***processors*** convert **output** chunks into instructions for motor ***actuators***[32] to act on CTM's outer world.

---

[30] Humans learn to play ping pong consciously. In a ping pong tournament however, a player must let the unconscious take over, must insist that the conscious get out of the way (see TableTennisCoaching.com ). In swimming, repetition gives one's unconscious an ṗ opportunity to improve one's stroke, but it doesn't enable a new stroke to be acquired. That requires conscious attention. For example, the dolphin kick is weird and unnatural, but since it works for dolphins, it makes sense to simulate it, and that is done consciously at first. The unconscious then optimizes the constants.

[31] Ears/sounds, eyes/sight, nose/smell, skin/touch, mouth/taste, … .

[32] Arms, hands, legs/motor actuators, mouth/vocal actuators, … .

### 2.1.7 Built-in Processor Properties

Now that we have established the basic CTM structure, we outline some important properties built into all processor at CTM's birth (**t = 0)**. These properties include *triggers* for what processors *must do* and come into play in *triggering* conscious awareness (section 2.3).

1. Each and every processor ***stores in its memory*** every chunk *it submits* to the competition and every chunk *it receives* by broadcast, forming two distinct lists ordered by time. We call these lists the processor's ***local dictionary*** and its ***global (Brainish) dictionary***, respectively.[33] Thus all processors have the same global dictionary.

2. ***Input*** (e.g., sensory) ***processors***, ***Output*** (e.g., motor) ***processors***, and ***Gauge processors*** (homeostats that monitor measurable quantities like fuel, temperature, general well-being, and so on) each have ***specialized functions built in***.

3. All processors (including the above specialized ones) have ***built-in learning algorithms*** that work (unconsciously) to predict, check, correct and improve that processor's prediction algorithms (section 2.1.8), to set values, to make chunks, to communicate with other processors, and so on. Importantly, a *Sleeping Experts Learning Algorithm* (necessarily modified for CTM) adjusts/corrects the value each processor gives its gists.[34] See (Blum A. , 1995) and (Blum, Hopcroft, & Kannan, 2015). [KM6]

---

[33] That's two chunks per tick. The decision was made to not store every chunk transmitted along links (section 2.1.5) because a processor can in general have almost **N** links, too much to store in a single tick or even a few ticks. A consequence of this decision is that if CTM is asked at some point why it did something it did, it can answer based on what it knows from broadcasted chunks, from submitted chunks, but not from chunks sent along links, which thwarts its ability to answer in full.

[34] Referring to the **"What's her name?"** scenario (footnote in section 2.1.2), recall that processor **p**' got its incorrect information quickly into STM while it took processor **p"** longer to get its correct information in. The *Sleeping Experts Algorithm* in **p'** caused **p'** to lower the importance it gives its information, while the *Sleeping Experts Algorithm* in **p"** caused **p"** to increase the importance it gives its information.

4. Each and every processor has a ***built-in preference for choosing greater (more positive) values*** over lesser (more negative) values.[35] By contrast, the Up-Tree competition has a preference for choosing greater intensities, or more generally, greater **f-values** (see section 6.2.)

5. For each processor, at any moment of time, some of its algorithms are critical and must be executed. The time left over is its ***free time***. On receipt of a broadcasted chunk **<p, t, g, v; aux>,** every processor is ***programmed to focus*** a fraction of its *free time* – a fraction proportional in part[36] to **|v|** - to ***deal with*** the chunk**.** If $\mathbf{v < 0}$, then the processor ***must spend*** that time on making **v** less negative.[37]

   **To *deal*** with the chunk *includes* trying to understand the gist, to carry out the task or solve the problem posed, and so forth. In order to understand it, processors ***may unpack*** the chunk (Formal Definition 7 in section 2.3.2.2).

6. Each processor has a built-in ***threshold*** parameter $\theta_t$**.** On receipt of a broadcasted chunk **<p, t, g, v; aux>**, if $|v| > \theta_t$, then the processor ***must unpack*** the broadcasted chunk. If in addition, $v < 0$, then the processor ***must spend*** *as much of its time as possible* on making **v** less negative. (This may force the processor to abandon some of its essential duties.)

### 2.1.8 Predictive Dynamics (Prediction + Testing + Feedback + Correction/Learning)

A major goal for CTM is to ensure that all predictions are as accurate as possible. For the most part this is done unconsciously (locally) within the individual processors themselves. Processors make predictions when they turn raw sensory inputs into chunks (a kind of analog to digital conversion[38]), when they put chunks in the competition, when they send information over links, and when they transform chunks into raw outputs (a kind of digital to analog conversion). Processors get feedback from broadcasts, from inputs to CTM, and through links. Learning/correcting then takes place within processors.

---

[35] This built-in *preference* in the processors (not in the competition function) for choosing positive over negative contributes to the drive for survival.

[36] "In part" depends on the priorities of the processor **p**.

[37] Note, we have not stipulated specifically what to do if a broadcasted chunk has high positive **v**. This is purposeful. As a consequence, except for other constraints (e.g., responding to tasks), positivity allows processors to use more of their free time as they choose. This is an example where positivity and negativity are not symmetric in the CTM.

[38] Digital signals are binary and have a few defined states, while analog signals have a theoretically infinite number of states.

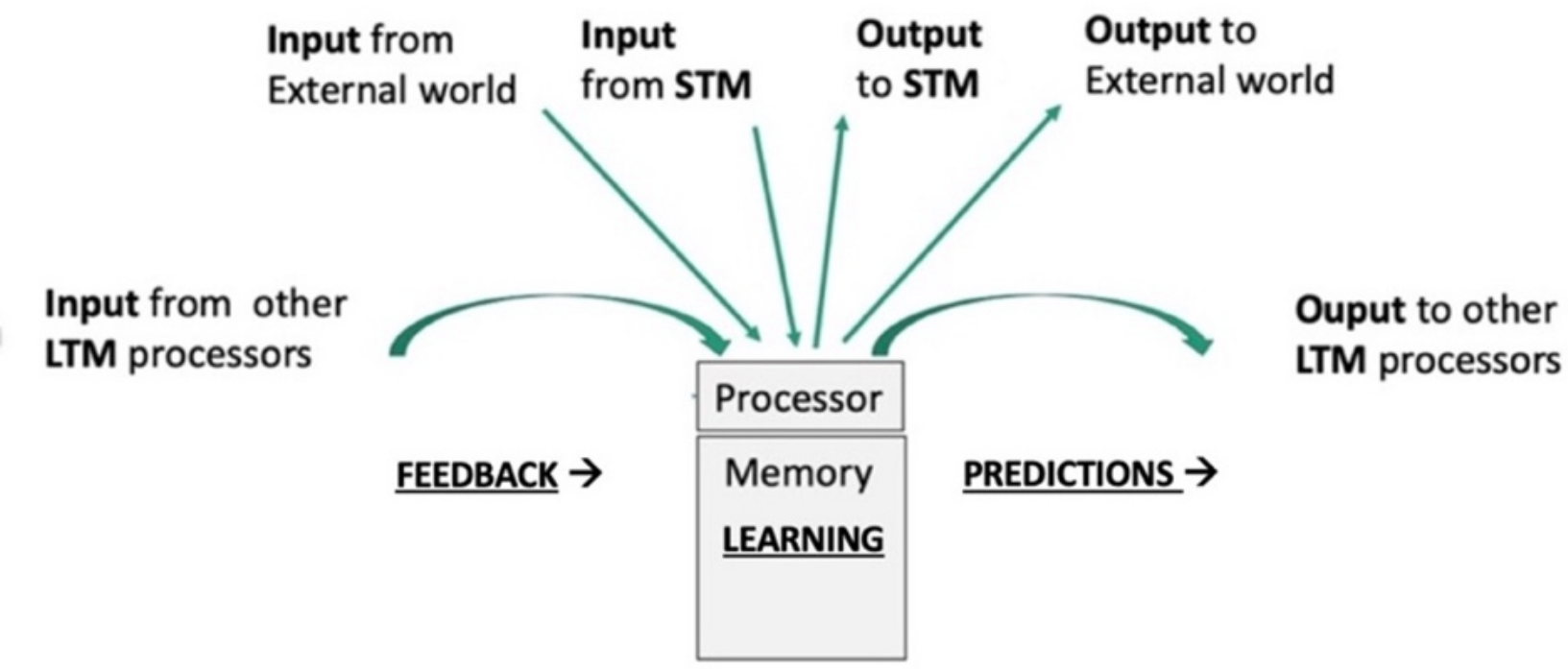

**Figure 3. Processors predict (out), get feedback (in), learn (within).**

Thus, we can already see some basic ***predictive dynamics*** (cycles of: prediction → testing → feedback → correction/learning) occurring locally and globally within CTM.

## 2.2 Conscious Attention in CTM and Axiom 1 (Global Broadcast Axiom)

> **Formal Definition 1**. We say CTM ***pays conscious attention*** to the winning chunk when this chunk is simultaneously received by all LTM processors.[39]

We state our first axiom, the ***Global Broadcast Axiom***.

---

[39] There is a delay of **h** clock ticks between when LTM processors submit their chunks to the competition for STM and when the winner appears in STM.

In one more clock tick, CTM pays conscious attention to the winner. This **h+1** delay is somewhat analogous to behavioral and brain studies going back to (Libet, 1985) that suggest a delay between when unconscious processors in our brains make decisions and when we become conscious of them. The CTM competition suggests, however, that while some processor proposed a decision that became conscious h+1 ticks later, others proposed decisions that did not win. So, in a real sense, the broadcast decision in CTM may *not* have been fully made.

**Axiom 1.** The simultaneous reception of a broadcasted chunk by all processors *evokes* in CTM a *unitary* experience.

Hence, *conscious attention* evokes a unitary experience in CTM.

We call a finite sequence of successively broadcasted chunks a ***stream of consciousness***, more specifically, a ***stream of conscious attention***.

Globally, the process of competition for STM, global broadcast to LTM, and consequent processor responses and direct communication via links is reminiscent of a process that Stanislas Dehaene and Jean-Pierre Changeux have called *ignition* (Dehaene & Changeux, 2005).

Our definition of *conscious attention* is related to what is often called *access consciousness* (Block, 1995).

But... for *subjective feelings of consciousness*, attention is not all you need. More is required.

## 2.3 Conscious Awareness: "What it is like to be" a CTM[40]

How might CTM, a formal machine model, experience feelings such as *pleasure* and *pain*? How might CTM experience the *subjective feelings of consciousness*?

In this section we look at how the *Model-of-the-World* (MotW) and *Brainish*, CTM's self-generated multimodal inner language, co-evolve to play an essential role in *"what it is like to be"* a CTM.

We begin with some preliminary concepts, informally, indicating (in section 2.3.2) how we formalize these concepts in CTM.

The ***Model-of-the-World processor*** (**MotWp**) is not actually a single processor. Its functionality and memory are distributed across all LTM processors. The MotWp builds models of CTM's inner and

[40] Thomas Nagel's "what it is like to be" (Nagel T. , 1974) is often taken to be the canonical "definition" of phenomenal or subjective consciousness.

outer worlds which also are distributed across all LTM processors. We call this *collection* of models, the ***Model-of-the-World*** (**MotW**).[41] [KM1][KM2][KM8]

The MotW is CTM's current and continuing view of CTM's (inner and outer) worlds. Specifically, the MotWp constructs chunks with *succinct **Brainish** descriptions* (gists) of ***referents*** in CTM's (inner and outer) worlds. When these chunks get broadcasted, they become ***sketches*** in the MotW. [KM1]

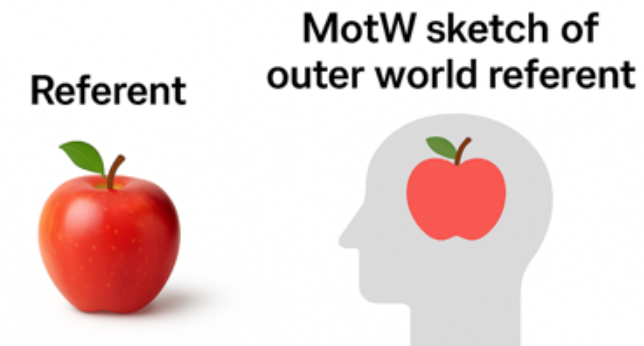

**Figure 4. MotW sketch of a referent.**

A ***referent*** might be a (red) apple or (red) rose in CTM's outer world, or a feeling (of pleasure), or a thought, idea, desire, concern, and so on in its inner world. Sketches of inner world referents are as important as sketches of outer world referents.

Sketches may be ***labeled*** with ***Brainish words*** such as (translated into English): BRIGHT_COLOR ∘ SWEET_TASTE ∘ CRUNCHY or BRIGHT_COLOR ∘ SWEET_SMELL ∘ SILKY_TOUCH or HAPPY_THOUGHT. The concatenator ∘ indicates the ***fusion*** of previously defined Brainish words. (See Formal Definition 4 in section 2.3.2.) [42]

These *labels* of sketches are Brainish words for what CTM "senses" or "feels" about those referents based on "lived" experiences.

---

[41] Although the ***Model-of-the-World*** is widely distributed, it is a "useful fiction" to think of it as a *single* (evolving) entity.

[42] A human dream frame typically has both outer and inner world labeled sketches. For example, a nightmare frame might contain a sketch of monsters under the bed labeled FEAR. The feeling in the dream is one of great fear.

The *labels* ***represent*** the ***qualia (quale) evoked*** when CTM became ***consciously aware*** of their ***meanings***. (See section 2.3.2 where these concepts are formalized.)

Over time and with experience, Brainish words, labels and sketches (Formal Definitions 4 and 5 in section 2.3.2) get introduced and become finer, richer and more nuanced.

In particular, the sketch "CTM" in the MotW, whose referent is CTM itself,[43] will develop from scratch and may eventually be labeled with SELF ∘ CONSCIOUS, etc. The sketch starts as an empty blob representing both the newborn CTM and the "world". In time the "CTM" sketch will become an entity distinct from the rest of the "world" (see sections 2.3.1.2 and 2.3.1.3).

CTM will pay ***conscious attention*** to MotW labeled sketches when broadcasted chunks referring to these *labeled sketches* are received by all processors. A ***unitary experience*** is ***evoked*** (**Axiom 1** in section 2.2). If in addition, all processors ***unpack*** the received chunks (Formal Definition 7 in section 2.3.2.2), CTM will become ***consciously aware*** of these labeled sketches (Formal Definition 9 in section 2.3.2.2) and they will ***evoke unique subjective experiences*** (**Axiom 2** in section 2.3.2.2). [KM3][KM11][KM15]

In this way, **MotW becomes the world that CTM *consciously* "sees", or more generally *consciously* "senses" or "knows".** [KM1] [KM3] [KM15]

### 2.3.1 Co-Evolution of Brainish and the Model-of-the-World (MotW)

The Model-of-the-World (MotW) and Brainish co-evolve during CTM's lifetime, starting at birth ($\mathbf{t = 0}$). To motivate our formalization (in section 2.3.2) of this co-evolution, we start here with some examples.

---

[43] We are often asked, isn't this process recursive? Doesn't the sketch of CTM have a sketch of CTM have a sketch of CTM, etc.? Yes, up to a point. But, at each iteration the current sketch is degraded, so the process rapidly ends up with the empty sketch.

The MotW in the newborn CTM (which could be a human infant if it has a CTM brain) contains just an empty blob at first representing both the newborn CTM and the “world”. This blob acquires CTM’s first Brainish name as follows:

Suppose the chunk **<$p_0$, 0, NIL, $v_0$; aux>** is CTM’s very first broadcasted chunk. Here **NIL** is a gist that represents the empty sequence. The pair **<$p_0$, 0>** is a pointer to that chunk, **<$p_0$, 0, NIL, $v_0$; aux>.** ***Brainish words*** will in general be such pointers.[44]

> **Formal Definition 2.** CTM’s very first broadcasted chunk **<$p_0$, 0, NIL, $v_0$; aux>** is its ***initial sketch*** in the MotW of both the “world” and the newborn CTM. The pointer **<$p_0$, 0>** to the sketch **<$p_0$, 0, NIL, $v_0$; aux>** is CTM’s ***primal Brainish word***, and it is the ***name*** of that initial sketch. The English translation of the Brainish word **<$p_0$, 0>** is **$BLOB_0$.**

In general, *sketches* will be broadcasted chunks **<p, t, gist, v; aux>** (originating from the MotWp) and *names of sketches* are pointers **<p, t>** to these chunks. (See Formal Definition 5 in section 2.3.2.) It is sometimes convenient to identify the *name of the sketch* with the *sketch* itself.[45]

Since processors store all broadcasted chunks (built-in property 1 in section 2.1.7) any LTM processor at any future time can ***unpack*** the word **<p, t>**, that is, look into its memory and *retrieve*

---

[44] Brainish *words* are reminiscent of what, in the AI/ML/neuroscience literature, are called *keys*; *chunks*, that words point to (the key’s *meaning*), are reminiscent of what are called *values* (Gershman, Fiete, & Irie, 2025).

[45] "You are sad," the Knight said in an anxious tone: "let me sing you a song to comfort you." … “The name of the song is called 'Haddocks' Eyes.'" "Oh, that's the name of the song, is it?" Alice said, trying to feel interested. "No, you don't understand," the Knight said, looking a little vexed. "That's what the name is called. The name really is 'The Aged Aged Man.'" "Then I ought to have said 'That's what the song is called'?" Alice corrected herself. "No, you oughtn't: that's quite another thing! The song is called 'Ways And Means': but that's only what it's called, you know!" "Well, what is the song, then? " said Alice, who was by this time completely bewildered. "I was coming to that," the Knight said. "The song really is 'A-sitting On A Gate': and the tune's my own invention." (Carroll, 1871)

and *examine (take note of)* the stored chunk, **<p, t, gist, v; aux>**.[46] (See Formal Definition 6 in section 2.3.2.2.)

We view the procedure in processor **$p_0$** that generated the broadcasted chunk **<$p_0$, 0, NIL, $v_0$; aux>** to be the part of the MotWp in **$p_0$**.[47] We view this chunk (sketch) which is now stored in every processor to be in the MotW.

Brainish words for pain and pleasure also develop early. We indicate here how these words come to be in CTM. We start with pain.

#### 2.3.1.1 Brainish Word for Primal PAIN

Consider again an infant CTM at the moment it is born. A processor that monitors the $O_2$ level, call it the **$O_2$_Gauge**[48], raises its growing concern for the lack of $O_2$ by submitting to the Up-Tree competition a sequence of chunks having ***negatively valenced values*** of ***increasingly high magnitude***. At some point, the $O_2$_Gauge processor's chunks get onto the stage (STM) and are broadcasted, their huge and growing negative value signaling an increasingly desperate "scream" for something ($O_2$). All processors hear this "scream".

Since every processor stores every chunk that ever got broadcast (built-in property 1 in section 2.1.7), every processor stores the $O_2$_Gauge processor's broadcasted chunks, **<$p_1$, $t_1$, $gist_1$, $v_1$; aux>**. Here **$p_1$** is the $O_2$_Gauge processor's address, **$t_1$** is (any) one of the times at which the $O_2$_Gauge processor screams for $O_2$, and **$gist_1$** (**= NIL**) has a high (and growing) negative value, **$v_1$**. The high (and growing negative value causes each processor to focus a fraction of its free time (proportional to **$|v_1|$**) to increase **$v_1$**, (make it less negative) something all processors will have been programmed to do (built-in properties 5 and 6 in section 2.1.7).

---

[46] Retrieving the stored chunk can be done quickly. That's because every processor stores every broadcasted chunk (built-in property 1), and the processor is a random-access machine. A Turing Machine could not do the lookup quickly enough.

[47] Recall, CTM's MotWp is distributed across all LTM processors.

[48] "$O_2$_Gauge" is a metaphorical name for a CTM gauge processor that, like all CTM gauges, has a built-in program to announce a great need. (See built-in properties in section 2.1.7.)

The processors have absolutely no idea what to do (as the infant was just born) to reduce that huge and growing |value|. They know only that they must do something. Processors that control CTM's actuators command them to do something, anything. The arms and legs flail. The infant pees and poos. The vocal actuator screams and cries. This last one works! The cry opens the "lungs" and the infant takes in its first breath. The $O_2$_Gauge processor's value gradually returns to normal.

The next time the infant CTM needs help, it finds that screaming and crying work again. In short order, the infant CTM learns that a good response to negativity (particularly intense negativity) of any kind is to scream and cry.

Recall that a chunk **$<p_1, t_1, NIL, v_1; aux>$** was a scream for help from the $O_2$_Gauge processor. Let **$<p_1, t_1>$** be a pointer to this chunk.

> **Formal Definition 3.** The Brainish word **$<p_1, t_1>$** is a ***primal Brainish word for pain***. Its English translation is **$PAIN_1$** (or PAIN_FROM_THE_LACK_OF_$O_2$).

Different Brainish words for pain will evolve over time, later pains building on earlier ones.

Now that CTM has these two Brainish words, **$<p_0, 0>$** (i.e., **$BLOB_0$**) and **$<p_1, t_i>$** (i.e., **$PAIN_1$**), more complex words and phrases can be generated. For example, stringing these two words together creates the ***Brainish phrase***, **$<p_0, 0><p_1, t_1>$**,[49] translating to the English phrase "**$BLOB_0$ is in $PAIN_1$**".

Now, some processor **$p_2$** may submit a chunk **$<p_2, t_2, g_2, v_2; aux>$** into the Up-Tree competition, where the gist **$g_2$** is the ***Brainish phrase***, **$<p_0, 0><p_1, t_1>$**. If **$|v_2|$** is large enough, the chunk may win the competition. Then the Brainish word **$<p_2, t_2>$**, a pointer to **$<p_2, t_2, g_2, v_2; aux>$**, will be a ***fused*** Brainish word.

The above indicates how the Brainish language starts to develop.

---

[49] As in the simple constructed language, Toki Pona (Lang, 2014), and in many languages, we are placing the adjective after the noun. However, we are not formally specifying a Brainish grammar. For the moment, a Brainish phrase is just a multi-set of Brainish words.

Other primal Brainish words quickly develop. For example, suppose **$p_3$** was the ***Vocal processor*** whose direct signals to the ***vocal actuator*** caused the newborn **CTM** to cry. When the processor learns that its signal produced the desired result, it submits a chunk **<$p_3$, $t_3$, $gist_3$, $v_3$; aux>** with **$gist_3$ = NIL** announcing this discovery to the competition with high enough **|$v_3$|** so that its chunk gets broadcast.

Then **<$p_3$, $t_3$>** becomes a primal Brainish word for **CRY**.

#### 2.3.1.2 Primal HUNGER, PLEASURE and SELF

Now suppose the infant CTM's *Fuel Gauge* is low. The *Fuel Gauge processor* creates a chunk with negative value proportional to the need. When the value is sufficiently negative, this chunk will (with high probability) win the competition and get globally broadcast. Again, the processors are "compelled", i.e. programmed, to do or try to do something to lower the |value|.

Having learned that crying is a good response to negativity, the CTM cries out causing the *Fuel Source* (mother) in the outside world to respond. As the fuel comes in, the Fuel Gauge will start submitting chunks with more positive values (... → -4 → -2 → ...).

Let **<$p_4$, $t_4$, $gist_4$, $v_4$; aux>** be a chunk that, on account of its extreme negative value **$v_4$**, is a scream to do something (to make **$v_4$** more positive). Here **$p_4$** is the Fuel Gauge processor's address, **$t_4$** is one of the times at which the chunk was broadcast, and **$gist_4$** could be anything, for example, **NIL**.

The *pointer* **<$p_4$, $t_4$>** to **<$p_4$, $t_4$, $gist_4$, $v_4$; aux>** is now a Brainish word that translates (we suggest) to the English **HUNGER_PAIN**. The Fuel Source sketch (of mother) in the MotW will eventually get labeled with (we suggest) **RELIEVES_HUNGER_PAIN** and **PLEASURE_PROVIDER**.

In this way, the newborn CTM learns (or increases its confidence) that **PLEASURE** relieves **PAIN**.[50]

---

[50] When a human mother gives a breast to her infant, the infant learns that the breast relieves the pain of hunger. The breast gets incorporated into the infant's MotW labeled with **RELIEVES_PAIN ∘ GIVES_PLEASURE**. Unless brought up in a dysfunctional household, the human infant learns that pleasure relieves pain and prefers pleasure over pain. This is one of many ways in which pain and pleasure are not symmetrical. This dynamic is reminiscent of the pleasure cycle

At this stage in CTM's early life, there is only one blob, named **$\mathbf{BLOB_0}$** in the MotW which represents the newborn CTM and the "world". With the introduction of the Fuel Source (mother), the blob will also represent mother. But soon enough, the infant CTM will discover that it and the Fuel Source are not one. The MotWp then separates the sketch named **$\mathbf{BLOB_0}$** into a sketch named **$\mathbf{CTM_0}$** and labeled **$\mathbf{SELF_0}$**, and a sketch named **FUEL-SOURCE**. (This is done by creating relevant chunks that get broadcast. See Formal Definition 5 in section 2.3.2.) [KM2]

#### 2.3.1.3 The Power of Thought and Sense of Self

The infant CTM's primal sense of self (**$\mathbf{SELF_0}$**) arose when it discovered that it was not the only entity in the world. In time, the infant CTM will discover that it can utilize the ***power of thought*** **(PoT)** to check what is **SELF** and what is not. [KM2].

To perform an action by the ***power of thought*** (or mental causation) is to perform the action in the MotW, then confirm that it actually got done in the world.

For example, at some point in time, the **MotW** will hold a rough sketch of the infant's left leg. When the infant **CTM** discovers it can move its left leg (an actuator) by the **power of thought**, the **MotWp** appends the label **SELF** to that sketch of the left leg. The scenario is as follows:

> The MotWp moves the left leg sketch in its MotW, which commands (via the Motor processor) the left leg to move. The MotWp detects that movement via CTM's sensors. *By repeating the action a number of times*, the infant CTM becomes ever more certain that it itself is responsible for moving the leg (that is, that this prediction is correct). Now convinced, the MotWp labels the left leg sketch with **SELF.** This Brainish label represents CTM's discovery that it can move its leg by the power of thought.[51] [KM13]

---

proposed by (Berridge & Kringelbach, 2015) consisting of three separate entities: wanting, liking, and learning.

[51] Certain pathologies will occur if a breakdown in CTM causes its MotWp to mislabel a sketch. For example, if the sketch of a leg gets mislabeled NOT-SELF, CTM might beg to get its leg amputated, even though the leg still functions properly. This would be an example of body integrity dysphoria (body integrity identity disorder) in CTM. Other pathologies due to faulty labeling in the MotW include: phantom limb syndrome (the sketch of an arm remains labeled

This explanation of the *power of thought* in labeling a limb as **SELF,** requires some qualification. In similar fashion, the newborn infant might easily surmise by the power of thought that its mother is SELF. After all, every time the newborn infant cries, its mother is activated. And so, in the same manner, the mother blob in the MotW is labeled **SELF**. But soon enough, the infant will discover the mother does not always respond, and the **SELF** label on mother will be removed.

### 2.3.2 Formalizing Brainish, MotW Labeled Sketches and Conscious Awareness

The primal Brainish words for pain and hunger, and early Brainish gists such as **BLOB$_0$ is in PAIN$_1$,** provide clues on how to *formally* define *Brainish words*, *gists*, and (*multimodal*) *fusion*.

#### 2.3.2.1 Brainish Words, Meanings, and Labeled Sketches

**Formal Definition 4.**

i. A ***primal Brainish word*** is a pointer **<p, t>** to a *broadcasted chunk* **<p, t, NIL, v; aux>** (where **NIL** denotes the empty sequence).

ii. A ***Brainish word*** is a primal Brainish word or a pointer <p, t> to a *broadcasted chunk* **<p, t, gist, v; aux>**. This chunk is the word's ***meaning***.

iii. A ***Brainish gist*** (***phrase***) is a multi-set of at most **8** *Brainish words*[52] (so a *word* is also a *gist*).

---

SELF after it is amputated), Cotard's syndrome (the sketch of CTM is labeled DEAD), paranoia (the sketch of CTM's best friend is labeled SPY), … . (There may be various causes for breakdowns in CTM, e.g., faulty sensors or sensory processors, faulty gauges, faults in the Up-Tree competition and/or broadcasting system, and so on.) [KM9]

[52] For Brainish words, we use **<p, t>** instead of the full chunk **<p, t, gist, v; aux>** because **<p, t>** is sufficiently short to appear 8 times in a gist while **<p, t, gist, v; aux>** is not short enough to appear even once. Given a word **<p, t>,** a processor can do a quick look up in its personal global memory for the associated **chunk**. Quick look ups are possible because the **CTM** processors are RAMS (Random Access Machines), not Turing Machines.

As commented earlier, we could require gists to be at most 2 or 3 Brainish words instead of 8, but for creating non-trivial examples, 8 is helpful (and the maximum for quick computations in the Up-Tree competition).

iv. If the ***Brainish word*** **<p, t>** points to **<p, t, gist, v; aux>,** we say the word **<p, t>** is a ***fusion*** of the words in the **gist**. Its ***meaning*** is a fusion of the meanings of the words in the gist.[53, 54]

Since all processors store all broadcasted chunks in their global dictionaries, when given a **word** each processor can each look up its ***meaning*** (i.e., *retrieve* the chunk it points to and *take note of* its contents). We call this ***unpacking*** the word. (See Formal Definition 6 in section 2.3.2.2.)

**Formal Definition 5.**

**i.** MotW ***sketches*** are broadcasted chunks that were submitted by the MotWp; ***names of sketches*** are pointers to those chunks.

ii. **Labels** (of sketches) are **words**, i.e., **NIL** or pointers to broadcasted chunks.

**iii.** Suppose **σ** is the *name of a sketch* (such as **<p, t>**) and **λ** is a *label* (such as **<p', t'>**). The Brainish (two-word) gist **σλ** denotes the English phrase **"the sketch named σ is labeled λ"** or more succinctly **"σ is labeled λ".** We call the gist **σλ,** a ***labeled sketch***.

Sketches and labels get refined over time.

### 2.3.2.2 Working Axiom 2 (Unpacking Axiom), Conscious Awareness, Qualia, and the Feeling of Consciousness

**Formal Definition 6.** A processor ***unpacks*** *a Brainish word* ⟨p, t⟩ by *retrieving* (from its global Brainish dictionary) the chunk to which ⟨p, t⟩ points (the word's *meaning*) and ***reading*** it. (*Reading* means accessing the chunk's components and making them available for further processing.) We consider the NIL word *unpacked*.

---

[53] As Brainish words evolve, CTM's Brainish language may develop a simple basic (creole-like) grammar. For information on such (simple creole-like) languages see: (McWhorter, 1998), (McWhorter, 2008), (Sigal, 2022) and (Bancu, et al., 2024).

[54] Trivial observation: If **<p, t, gist, v; aux>** is a broadcasted chunk and **<p', t'>** is a word in its **gist,** then **t' < t.**

**Formal Definition 7.**

i. We say a processor ***unpacks*** a chunk **<p, t, g, v; aux>** whenever it ***examines/takes note of*** its components and ***unpacks*** all words **<p', t'>** appearing in gist **g.**
ii. We say **CTM** ***unpacks*** a broadcasted chunk **<p, t, g, v; aux>** whenever all processors simultaneously (in the same tick) ***unpack*** that chunk.[55]

This is the *first level* of unpacking by CTM. A *deeper unpacking* of **<p, t, g, v; aux>** does not require simultaneity. It can be done by a single processor involving its personal *priorities*.

The chunk's *value* **v** signals the *importance* and *valence* that the processor **p** gives the gist. (The *auxiliary information* **aux** provides "global context". (See section 2.1.3 and section 6.2 in the Appendix for a discussion of auxiliary information.) By built-in properties 5 and 6 (section 2.1.7), if **<p, t, g, v; aux>** is a broadcasted chunk and $|\mathbf{v}| > \theta_t$, then all processors *must unpack* the chunk and spend a fraction of their time *dealing with* gist **g.** If $\mathbf{v} < \mathbf{0}$, then all processors ***must spend*** *as much of their time as possible* on making **v** less negative. (This may force the processor to abandon some of its essential duties.)

We now state our **Working Axiom 2,** the **unpacking** ***Axiom*****,** for ***subjective experiences*** in CTM.

**Working Axiom 2.** When **CTM** ***unpacks*** a chunk **<p, t, g, v; aux>** it ***evokes*** in **CTM** a ***unique subjective experience.*** The *intensity* and *valence* of that *evoked experience* is given by **v** modulated by the "global context" given by **aux.**

Specifically:

i. If **g** is **NIL**, when **CTM** ***unpacks*** a chunk **<p, t, NIL, v; aux>** it ***evokes*** in CTM a ***unique subjective experience*** and that experience, whatever it might be, is a ***primal subjective***

[55] We assume CTM can *unpack* a word in **.1** milliseconds, 8 words in at most **1** millisecond, and *unpack* a chunk in **2** milliseconds.

***experience***.[56] The *intensity* and *valence* of that *evoked experience* is given by **v** modulated by the "global context" is given by **aux**.

ii. If gist **g** = $\langle p_1, t_1\rangle \ldots \langle p_k, t_k\rangle$, then when CTM ***unpacks*** a chunk **<p, t, g, v; aux>** it ***evokes a unique subjective experience*** in CTM that is a ***fusion*** of the ***subjective experiences evoked*** by the ***meanings*** of the words in g. The *intensity* and *valence* of that *evoked experience* is given by **v** modulated by the "global context" given by **aux**.

**Formal Definition 8.** *Conscious awareness* in CTM arises when CTM pays *conscious attention* to a chunk **<p, t, g, v; aux>** and *unpacks* it.

**THEOREM.** When CTM becomes *consciously aware* of a broadcasted chunk **<p, t, g, v; aux >**, a *unique unitary subjective experience is evoked*. This subjective experience is a ***fusion*** of the ***subjective experiences evoked*** by the *meanings* of the words in **g**. The *intensity* and *valence* of the *evoked experience* is given by **v** (modulated by "global context" **aux**).

**Proof.** CTM becoming *consciously aware* of a broadcasted chunk **<p, t, g, v; aux >** means that CTM pays *conscious attention* to it (i.e., all processors simultaneously receive that broadcasted chunk) and *unpacks* it. The conclusion follows from **Axiom 1** and **Axiom 2.**

Suppose the gist **σλ** is a *labeled sketch* where **σ** is the *name* of a MotW *sketch* and **λ** one of its *labels*. By the above **Theorem**, a *fused subjective experience is evoked* when CTM becomes *consciously aware* of a chunk **<p, t, σλ, v; aux>**.

Thus, for example, when CTM becomes ***consciously aware*** of a (chunk with) MotW labeled sketch of the referent rose with fused label BRIGHT_RED ∘ SWEET_SMELL ∘ SILKY_TOUCH, it "sees" the bright red color, "smells" the sweet odor, and "feels" the silky petals of the rose. That is, those subjective experiences are evoked.

We now consider two related notions of *qualia*, one we call *platonic*, the other *enacted*. *Platonic qualia* refer to *subjective experiences* "evoked" by labels (i.e., *evoked* by their meanings). Labels might be RED or SWEET or SOFT. *Enacted qualia* refer to *subjective experiences* "evoked" by labeled sketches. In CTM, the qualia evoked by being consciously aware of a "red rose" are different from that evoked by a "red fire engine".[57]

---

[56] The bottom-most turtle.

As CTM becomes consciously aware of more and more broadcasted chunks, some processor may call attention to this situation. Specifically, some processor may submit a chunk to the Up-Tree competition asserting that CTM is becoming ever more consciously aware. If this chunk wins the competition and hence gets globally broadcast, we say the CTM has perceived – or has become perceptive of – its conscious awareness. If in addition, this broadcasted chunk is simultaneously unpacked by all processors, we say the CTM becomes consciously aware of its conscious awareness. This is *a higher order of conscious awareness* in CTM and the MotWp will now label CTM in the MotW as CONSCIOUSLY_AWARE or simply CONSCIOUS.

Now suppose the sketch of CTM has also been labeled SELF. When CTM becomes *consciously aware* of a (chunk with) labeled sketch <CTM><SELF ∘ CONSCIOUS>,[58] CTM *will experience* (according to **Axiom 2**) *a fusion* of the *subjective experiences evoked* by the meanings of the words in this gist. CTM will *feel* itself conscious. [KM3] [KM9]

#### 2.3.2.3 The *Experience* of Pain

Pain is a particularly good exemplar of the Hard Problem as it is hard to imagine that a robot can truly *suffer* from pain, not just *simulate* suffering. Here we look into how CTM *experiences* the feeling of pain.

It follows from **Axiom 2** that becoming *consciously aware* of a chunk **<p, t, g, v; aux>,** whose gist contains a Brainish word for pain, **<p', t'>,** will *evoke* a fused experience that includes the feeling of pain **<p’, t’>** previously experienced by CTM. The label **<p, t>** represents a fused *platonic quale* (*qualia*).

---

[57] The CTM definition of enacted qualia is in the spirit of (Merleau-Ponty, 1945; 1962, English translation, Colin Smith) and (Varela F. , 1996).

[58] That is, all processors simultaneously receive the broadcasted labeled chunk with gist **<CTM><SELF ∘ CONSCIOUS>** and simultaneously unpack this gist.

Different kinds of pain evolve over time. Later pains build on earlier ones. For example, when the infant CTM becomes a toddler and first skins its knee, its MotW will contain a sketch of a bloody knee. As soon as the toddler CTM becomes consciously aware of this new pain, it *experiences* a fused feeling of earlier pains (**Axiom 2**). A more nuanced Brainish label is created: SKINNED-KNEE_PAIN.

When a sketch of a bloody skinned knee is labeled SELF but not PAIN, or PAIN but not SELF, CTM does not consciously experience it. That is, in the case of SELF but not PAIN, CTM does not become consciously aware that itSELF has pain; in the case of PAIN but not SELF, the pain does not belong to itSELF. In either case, we say CTM has *pain asymbolia*.[59] [KM9] [KM14]

The distinction between *pain-knowledge* and *pain-suffering* is important when considering the *behavioral aspects* of experiencing pain. *Pain-knowledge* refers to *paying* attention to the existence of pain without necessarily experiencing the suffering. *Pain-suffering* refers to both *pain-knowledge* and the *subjective experience* of pain. Pain-suffering serves as a motivator for responding appropriately to the pain and its causes. This follows from the built-in properties of CTM:

> Built into each and every LTM processor is an internal command that, when CTM becomes ***consciously aware*** of a chunk with high |value| and negative valence, causes them to spend *all* of their free time to do something to lower that |value|. (See built-in properties 5 and 6 in section 2.1.7.) Thus, each processor is “motivated” to do something to “solve the problem”. The experience of pain is thus exacerbated by this internal command. That is, forcing all LTM processors to spend more time on reducing the pain and less on whatever else they "want" to do, is experienced as painful.

[59] A person who has pain and *knows* everything about it but lacks the *feeling* of agony has *pain asymbolia.* Such a person is not motivated to respond normally to pain. Children born with pain asymbolia rarely live past the age of three. The experience of pain, whether physical or emotional, serves as a motivator for responding appropriately to the pain. See (Grahek, 2001; 2007), (Klein, 2015), and (Gerrans, 2024).

## 2.4 CTM as a Framework for Artificial General Intelligence (AGI)

Before indicating CTM's alignment with a number of major theories of consciousness, we remark on CTM's potential to serve as a *framework* for constructing an Artificial General Intelligence (AGI). This is a result of CTM's global architecture (kindred to arguments made for global latent workspace by (VanRullen & Kanai, 2021)) and, at the same time, the result of an essential difference between CTM and Baars' global workspace, CTM has *no* Central Executive. This is a feature, not a bug:

> **CTM's competition process enables it to engage processors to solve problems, even though CTM does not know which of its processors might have the interest, expertise, or time to work on them.** (Blum & Blum, 2023)**.** [KM14].

Specifically, if CTM, meaning one (or more) of its LTM processors, has a problem it needs solved, it (the processor) can submit the problem to the competition as a chunk with high |value| giving it a chance to win and be globally broadcast to all processors. Processors with the interest, expertise and time to work on the problem can respond with appropriately |value|-ed chunks.[60] In this way, ideas from unexpected sources may contribute to solving the problem, and useful collaborations can emerge. [KM14]

A Central Executive would have to know which processors have the inclination, expertise, and resources to solve problems as they arise, but Baars' does not say how the Central Executive could have that. [KM14]

---

[60] If CTM's competition is based on the simple **f-value** of intensity (see Appendix, section 6.2), the processor can submit the problem to the competition with |value| = intensity of the currently broadcasted chunk, which is the sum of all |values| of all submitted chunks. In that case, the chunk with that high |value| will win, with probability ≥ 1/2 and become globally broadcast to all processors.

We predict, in fact, that a Central Executive is not needed for consciousness or for general intelligence.

## 3 CTM Aligns with Underlying Principles of Major Theories of Consciousness

We have presented an Overview of the CTM model. Now we indicate how, at a high level, the model naturally aligns with and integrates underlying principles of major theories of consciousness, further supporting our view that CTM provides a *framework* for building a conscious machine. We start with theories that have greatest alignment.

### 1. Global Workspace (GW)/Global Neuronal Workspace (GNW)

CTM aligns broadly with the architectural and global broadcasting features of the *global workspace* (GW) theory of consciousness (Baars, Bernard J., 1997), and at a high level with the *global neuronal workspace* theory of consciousness of neuroscientists Stanislas Dehaene, Jean-Pierre Changeux (Dehaene & Changeux, 2005), (Dehaene S. , 2014), and others.[61]

However, CTM differs from GW in significant ways. For example: CTM's competition for global broadcasting is formally defined; CTM constructs world models; and CTM purposely has no Central Executive.

### 2. Attention Schema Theory (AST) and the Self-Organizing Metarepresentational Account (SOMA)

CTM's ability to construct and utilize models of CTM's worlds (inner and outer), and the key role they play in CTM's *conscious awareness*, align closely with neuroscientist Michael Graziano's *attention schema theory* (AST) of consciousness (Graziano, Guterstam, Bio, & Wilterson, 2020). AST proposes that the brain is an information processing machine that constructs a simplified model of attention, just as it constructs a simplified model of the body, the Body Schema. According to AST, this Attention Schema provides a sufficiently adequate description of what it (the brain) is attending to for it to conclude that it is "aware".

---

[61] Additional references for GNW include: (Dehaene & Naccache, 2001), (Sergent & Dehaene, 2005), (Dehaene & Changeux, 2011), and (Mashour, Roelfsema, Changeux, & Dehaene, 2020).

The constructive co-evolution of Brainish and the MotW in CTM aligns in principle with the self-organizing metarepresentational account (SOMA) according to which "consciousness is something that the brain learns to do." (Cleeremans A. , 2011) and (Cleeremans, et al., 2020)

### 3. Predictive Processing (PP)

Predictive processing (correctly) asserts that the brain is constantly inferring, correcting and updating its predictions, generally based on motor outputs and sensory inputs. CTM's *predictive dynamics* (cycles of prediction, testing, feedback, and learning/correcting), locally and globally, align with various incarnations of *predictive processing* (von Helmholtz, 1866; 1962), (Friston K. , 2010), (Cleeremans A. , 2014), (Clark A. , 2015), (Hohwy & Seth, 2020), and others.[62]

### 4. Embodied Embedded Enactive Extended Mind (EEEE Mind)

CTM's ability to construct (and utilize) models of its worlds containing rich Brainish labeled sketches (leading to its *feelings of consciousness*) derives in part from its embodied, embedded, enactive, and extended (EEEE) mind.

This aligns with the "4E" view that consciousness, like cognition (Carney, 2020), involves more than brain function (Rowlands, 2010). For consciousness, the interconnected 4E's are:

- *Embodied*: Incorporating relations with the entity's *body parts* and *processes* is essential for phenomenal consciousness. See, (Damasio, 1994), (Edelman, 2006) and (Shanahan, 2005).[63]
- *Embedded,* and *Enactive*: Being *embedded in the outer world* and *enacting*/interreacting with it, thus affecting the world and creating experiences, is necessary for phenomenal consciousness. See, (Maturana & Varela, 1972), (Maturana & Varela, 1980), (Varela, Thompson, & Rosch, 1991), (Thompson, 2007), and (Clark A. , 2008).

[62] Other references include: (McClelland & Rumelhart, 1981), (Lee & Mumford, 2003), (Friston K. , 2005), (Seth, The Cybernetic Bayesian Brain - From Interoceptive Inference to Sensorimotor Contingencies, 2015), (Miller, Clark, & Schlicht, 2022).

[63] We note that here Shanahan views the global workspace as key to access consciousness, but that phenomenal consciousness requires, in addition, embodiment.

- *Extended*: Consciousness is further enhanced by the entity having access to considerable external resources (such as libraries, Google, ChatGPT, Mathematica, ...). See, (Clark & Chalmers, 1998).

CTM is *embedded* in its outer world and, through its *embodied* actuators, can *enact* in this world, thus influencing what it senses and experiences. CTM's "mind" is *extended* by information it gets from resources in its outer world, and from embedded (or linked) off-the-shelf processors.

### 5. Integrated Information Theory (IIT)

Integrated Information Theory, the theory of consciousness developed by Giulio Tononi (Tononi, 2004), and supported by Koch (Tononi & Koch, 2015), proposes a measure of consciousness called Phi that, in essence, measures the amount of feedback and interconnectedness in a system. CTM's extensive feedback (its predictive dynamics, globally and locally) and its interconnectedness (global broadcasts and its fused multimodal Brainish gists) contribute to a high Phi.

### 6. Evolutionary Theories of Consciousness

CTM aligns with aspects of *evolutionary theories* of consciousness:

Oryan Zacks and Eva Jablonka provide evidence for the evolutionary development of a modified *global neuronal workspace* in vertebrates (Zacks & Jablonka, 2023) reinforcing our assertion that an AI with a global workspace architecture could possess access consciousness.[64]

In *Sentience*, Nicholas Humphrey presents an evolutionary argument for the development of *phenomenal consciousness* in warm-blooded animals (Humphrey, 2023). In "The Road Taken" (Chapter 12 of *Sentience*), Humphrey spins a "developing narrative", starting with "a primitive amoeba-like animal floating in the ancient seas. Stuff happens. …" The resulting story provides a roadmap for how an entity might create world models and a sense of self. Indeed, this roadmap closely parallels the way CTM's world models evolve, and how CTM develops its sense of self and (subjective) conscious awareness.[65]

---

[64] See (Ginsburg & Jablonka, 2019) for an extensive treatise on the evolutionary development of consciousness and their Unlimited Associative Learning (UAL) theory of consciousness.

Thus, while the former evolutionary theory (Zacks & Jablonka, 2023) aligns with CTM's built-in GW architecture, the latter theory (Humphrey, 2023) aligns with CTM's development of subjective consciousness over time.

### 7. Extended Reticulothalamic Activating System (ERTAS) + Free Energy Principle (FEP)

At a high level, CTM aligns with ERTAS + FEP:

In *The Hidden Spring* (Solms M. , 2021), Marc Solms proposes that the source of consciousness is the arousal processes in the upper brain stem. More generally, Solms cites the Extended Reticulothalamic Activating System (ERTAS) as the generator of feelings and affects, enabling consciousness. Although GW models generally consider processors as performing cortical functions, CTM goes beyond that. There is nothing to preclude CTM from having Gauge processors that function as ERTAS.

In "The Hard Problem of Consciousness and the Free Energy Principle" (Solms M. , 2019), Solms states that the "system must incorporate a *model of the world*, which then becomes *the basis upon which it acts*".[66] Similarly, *conscious awareness* in CTM of its MotW broadcasted chunks becomes "the basis upon which it [CTM] acts".

(Solms M. , 2019) posits that "affective qualia" are the result of homeostasis. "Deviation away from a homeostatic settling point (increasing uncertainty) is felt as unpleasure, and returning toward it (decreasing uncertainty) is felt as pleasure". Our discussions of the role of CTM's Gauge processors and of pain and pleasure (in sections 2.3.1.1 and 2.3.1.2, and in (Blum & Blum, 2021)) align with this view.[67]

---

[65] We claim Humphrey actually gives a road map for how an entity, warm blooded or not, might create world models and sense of self. As an exercise, we have re-written part of Chapter 12 (Humphrey, 2023) , "The Road Taken", from the perspective of CTM and sent a copy to Humphrey. His reply, "It would be great if we could meld these theories." (Personal communication with Nick Humphrey, Oct 9, 2023.)

[66] Italics here ours.

According to (Solms & Friston, 2018), homeostasis arises by a system resisting entropy, i.e., minimizing free energy. This is enabled by a *Markov blanket* (containing the system's input sensors and output actuators) that insulates the internal system from its outer world. In CTM, predictive dynamics (cycles of prediction, testing, feedback and correcting/learning) works to reduce prediction errors, an analogue to minimizing free energy.

Evocatively, the well-known Friston diagram (Parr, Da Costa, & Friston, 2019) rotated 90 degrees clockwise, with Markov blanket separating internal and external states, is clearly realized in CTM:

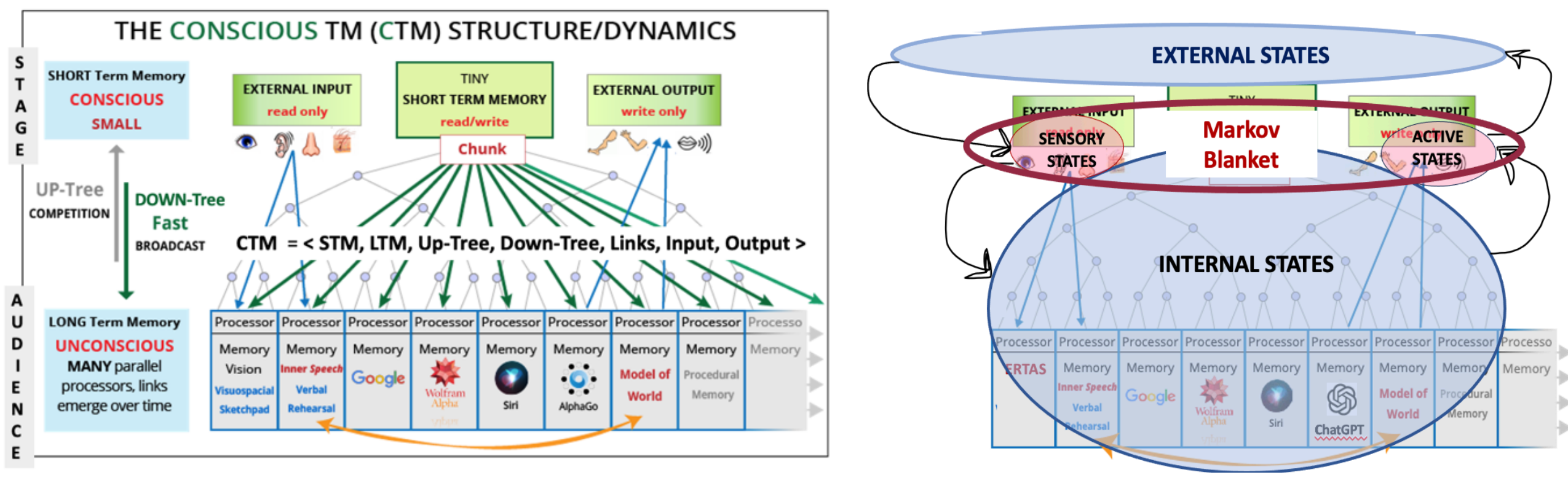

**Figure 5. Friston diagram rotated clockwise 90 degrees and superimposed on CTM.**

[67] Kevin Mitchell points out (personal communication) that another important point from Solms is that "the ascending homeostatic signals, which track different needs, must be valenced but also must have some distinguishing "qualities" so that when they are submitted to central decision-making units, the sources of the signals can be kept track of - so the organism doesn't mistake feeling thirsty for feeling tired (both of which feel BAD)".

In CTM, the fused Brainish gist FEELING_BAD ○ THIRSTY is different than the fused Brainish gist FEELING_BAD ○ TIRED. (Different Gauge processors were involved in generating these feelings and their traces are embedded in these feelings.) No need for a central decision-maker.

## 8. Comparison with Other Theories and Perspectives on Consciousness

Our view is that phenomenal or subjective consciousness is a consequence of (what is called) computational functionalism. Other researchers, including those mentioned above, tend to agree, at least to some extent. Features of CTM align with features and arguments they propose.

Lionel Naccache gives examples of how access consciousness can account for phenomenal consciousness (Naccache, 2018).

The co-evolution of Brainish and the Model-of-the-World[68] in CTM to produce subjective consciousness aligns with Axel Cleeremans et. al. proposal that phenomenal experience should be viewed "as the product of active, plasticity-driven mechanisms through which the brain learns to redescribe its own activity to itself." It is "not only shaped by learning, but its very occurrence depends on it." (Cleeremans, et al., 2020)

Our views on machine consciousness and AGI are close to (VanRullen & Kanai, 2021). We also see connections with Murray Shanahan's view on embodiment and the inner life (Shanahan, 2010). We see a kinship between the **CTM** and the self-aware robots developed by (Chella, Pipitone, Morin, & Racy, 2020). Leslie Valiant proposes a "computational model to describe the cognitive capability that makes humans unique among existing biological species on Earth". We see his model's employment of a "*Mind's Eye*, where descriptions of multiple objects and their relationships can be processed" (Valiant, 2024) is remarkably similar to CTM's employment of a Model-of-the-World for its conscious awareness.

We have not explored how CTM might align or not with higher order theories (HOTs) of consciousness. HOTs claim to take a middle ground between standard GWT and early sensory theories (Brown, Lau, & LeDoux, 2019). (LeDoux & Brown, 2017) argue "that the brain mechanisms that give rise to conscious emotional feelings are not fundamentally different from those that give rise to perceptual conscious experiences." We do see some connection between

[68] World models arise in other contexts, e.g., in the science of the brain and mind (Duncan, 2025). Here as in CTM, "the agent looks ahead to consider the future consequence of its choice." World models also arise in cognitive science, e.g., in analyzing what LLMs know (Yildirim & & Paul, 2024). We thank the anonymous referee for the latter reference.

HOTs specification of higher order representations (HORs) and CTM creating Brainish labeled sketches in its Model-of-the-World. Likewise, we see some connection between HOTs feature of *introspection* and the role thatunpacking plays in CTM's subjective experiences (Formal Definitions 7ii in section 2.3.2.2). As CTM becomes increasingly consciously aware it becomes consciously aware of its conscious awareness. We have yet to explore how *levels ofunpacking* in CTM might lead to higher *levels of subjective consciousness*.[69]

Philosophically, we align with much of Daniel Dennett's functionalist perspective (Dennett D. C., 1991). CTM's competition for STM leading to global broadcast and attention aligns with Dennett's "multiple-drafts" competing (Dennett D. , 2001) for "fame in the brain". Along with Dennett, we do not see the *explanatory gap* (Levine, 1983) as insurmountable. On the contrary, we see CTM as helping to explain the feeling of "what it is like" (Nagel T. , 1974). CTM, like Attention Schema Theory (AST), appears to embody and substantiate *illusionist* notions of consciousness proposed by Dennett (Dennett D. C., 2019) and Keith Frankish (Frankish K. , 2016).[70]

However, we are not claiming that any of these researchers also view that artificial consciousness is inevitable, even possible. For example, Anil Seth (Seth, 2025) and Nick Humphrey (Humphrey, 2023) view that consciousness requires living organisms.

---

[69] Relevant papers on levels and degrees of consciousness are (Farisco, Evers, & Changeux, 2024) and (Lee A. Y., 2023). We thank the anonymous reviewer for the latter.

[70] Saying that the feeling of consciousness is an illusion does not deny the existence of that feeling. It's just not what you think it is. As a familiar example, the fact that a movie is made up of (many) discrete still images does not affect the feeling of continuity one gets from viewing it. The feeling of continuity is an illusion.

In this regard, CTM and AST both align with philosophers Daniel Dennett's and Keith Frankish's views that consciousness is a kind of "illusion", which can be understood by such of their explications as: Consciousness is a "user illusion", the brain's interface with itself. (Dennett D. C., 1991); "Illusionism's a view about what experiences, perceptions, interpretations illusions *really are*. The illusion is that they are something they're not…" and "I believe that feelings are real, but I think we have mistaken ideas about what they are." (Frankish K. , 2023).

## 4 Summary and Conclusions

The Theoretical Computer Science (TCS) perspective has influenced the design and precise formal definitions of a simple formal machine model of consciousness, the CTM, and the conclusions we have drawn from the model.

Important features of CTM include:

- a formal competition leading to a global broadcast of the winner,
- a self-generated internal multimodal language (Brainish),
- interactions with the outer world via input sensors and output actuators,
- ability to construct models of its (inner and outer) worlds,
- cycles of prediction, feedback and learning, that constantly unconsciously (and consciously) update its algorithms,

all while operating under resource limitations (time and space).

Although CTM is inspired by the simplicity of Turing's formal model of computation and the insight of Baars' global workspace (GW) architecture, our formalization is neither a Turing Machine nor a standard GW model. Its *consciousness* (access and phenomenal) depends on what's under the hood, and its interaction with its worlds. In other words, what matters is not just the result of computation, but also *how* that computation was done.

In developing this theory, we were motivated to construct an "Occam's razor" **simple model** that addresses and provides insights into consciousness. We tried and discarded many variants of the CTM model because they grew in complexity without end. **The probabilistic CTM and only the probabilistic CTM has withstood our tests for simplicity and explanatory power.** This provides a measure of confidence that the CTM model is on the right track.

In an earlier paper (Blum & Blum, 2022), we provided examples of how CTM could exhibit phenomena associated with consciousness (e.g., blindsight, inattentive blindness, change blindness, and dreams) and related topics in ways that agree at a high level with cognitive neuroscience. Although our TCS approach differs from empirical science, we view its alignment at a high level with findings from empirical science as providing an additional measure of confidence. To the extent that CTM theory can predict other phenomena, or provide ways to think about questions related to consciousness, its perspective helps provide understanding.

The CTM model indicates how subjective experiences of pain and pleasure can be experienced by a machine. A major goal of the theory is to provide a fuller explanation and understanding of the hard problem of pain and pleasure.

As an example of a prediction from CTM, the theory proposes the existence of two different kinds of pain asymbolia (section 2.3.2.3.) One kind occurs when the sketch of the CTM in the MotW is labeled SELF but not PAIN, the other when labeled PAIN but not SELF. The SELF but not PAIN (i.e., sensory deprivation) explanation (Grahek, 2001; 2007) is standard. The PAIN but not SELF explanation was proposed by (Klein, 2015). (Sierra, 2009) quotes a typical pain asymbolic's explanation: "When a part of my body hurts, I feel so detached from the pain that it feels as if it were somebody else's pain." In an excellent and recent review, (Gerrans, 2024) points out that each of these two different explanations has a problem, without noting the possibility that there might be two different kinds of pain asymbolia which is what the CTM model predicts.

As an example of how CTM helps think us about other questions related to consciousness, here is a problem posed by (Zadra & Stickgold, 2021), p. 267:

> "Recall something that happened to you yesterday and think about it for a second. Got it? Well, as scientists, we don't have much of an idea of how your brain did that, how it searched yesterday's memories, selected just one of them, and found the associations that defined what it meant to you. And we know nothing about how you became conscious of any of this information."

Here is a CTM response to the request "Recall something that happened yesterday and think about it for a second":

> Different processors recall different things that happened yesterday and put those in the competition. One of them wins and its recollection gets broadcast. CTM *pays conscious attention* to whatever got broadcast. The prompt, "think about it for a second", will motivate processors to *unpack* the broadcast. As a consequence, CTM will experience subjective feelings related to the recalled memory, i.e., CTM will become *consciously aware* of what it recalls happened yesterday.

We see this explanation as providing a high-level understanding of how humans might respond, and become conscious of memories.

CTM is not a model of the (human or animal) brain, nor is it intended to be. It is a simple formal *machine model* of consciousness. Nevertheless, at a high level, CTM can exhibit phenomena associated with human consciousness (blindsight, inattentional blindness, change blindness, body integrity identity disorder, phantom limb syndrome, …), and aligns with and integrates those key features from main theories of consciousness that are considered essential for human and animal consciousness. The CTM model demonstrates the compatibility of those theories.

The fact that CTM is clearly buildable, and arguably a basis for consciousness, supports (the credibility of) our claim that *a conscious AI is inevitable.*

Finally, the development of CTM is a work in progress. More will appear in the upcoming monograph (Blum, Blum, & Blum, 2026). We view this paper as an extended outline for the monograph.

Our goal is to explore the model as it stands, determine the good and the bad of it, and make no (unnecessary) changes to it.

## 5 Acknowledgements

We are grateful to Michael Xuan for his enormous encouragement, and to UniDT for their long-term support. We appreciate numerous discussions with and helpful critiques from friends and colleagues: Johannes Kleiner, Wanja Wiese, Ron Rivest, Alvaro Velasquez, Sergio Frumento, Aran Nayebi, Martin Korth, Judy Roitman. We thank Kevin Mitchell for posing his enormously sensible questions "that a theory of consciousness should be able to encompass" and his subsequent perceptive discussion with us. We thank the anonymous reviewer for insightful comments and thoughtful suggestions.

# 6 Appendix

## 6.1 A Brief History and Description of the TCS Approach to Computation

The Theoretical Computer Science approach to computation starts with Alan Turing in the 1930's and focuses on the question, "What is computable (decidable) and what is not?" (Turing, 1937). Turing defined a simple formal model of computation, which we now call the Turing Machine (TM) and *defined* a function to be computable if and only if it can be realized as the input-output map of a TM. The formal definition of a TM (program) also provides a formal definition of the informal concept of algorithm.

Using his model, Turing proved properties (theorems) of computable functions, including the existence of universal computable functions (universal Turing Machines) and the fact that some functions, though definable, are not computable. The former foresees the realization of general purpose programmable computers; the latter that some problems cannot be decided even by the most powerful computers. For example, Turing shows there is no Turing Machine (Turing computable function) that given the description of a TM **M** and an input **x**, outputs **1** if **M** on input **x** (eventually) halts, and **0** if not. This is known as the "halting problem" and is equivalent to Gödel's theorem on the undecidability of arithmetic.

But why should we believe the Church-Turing Thesis, suggested first in (Turing, 1937), that the TM embodies the informal notion of computability (decidability)? That's because each of a great many very different independently defined models of discrete computation, including TMs and Alonzo Church's *effective calculability* (Church, 1936), define exactly the same class of functions, the computable functions (also called the *recursive functions*). In programming parlance, all sufficiently powerful practical programming languages are equivalent in that anyone can simulate (be compiled into) any other. The ensuing mathematical theory is generally called the Theory of Computation (TOC).

In the 1960's, with the wider accessibility of computers, newly minted theoretical computer scientists such as Jack Edmonds (Edmonds, 1965) and Richard Karp (Karp, 1972), pointed out that resources matter. Certain problems that in principle were decidable/computable, were seemingly intractable given feasible time and space resources. Even more, intractability seemed to be an intrinsic property of the problem, not the method of solution or the implementing machine.

The ensuing sub-theory of TOC, which introduces resource constraints into what is or is not *efficiently* computable, is called Theoretical Computer Science (TCS).

TCS focuses on many problems that arise when resource limitations are taken into account, including the question, "What is or is not computable (decidable) given limited resources?" A key problem here is the deceivingly simple "SAT problem": Given a boolean formula $\mathcal{F}$, is it satisfiable, meaning "Is there a truth assignment to its variables that makes formula $\mathcal{F}$ true?" This problem is decidable. Here is a decision procedure: Given a boolean formula $\mathcal{F}$ with **n** variables, systematically check to see if any of the $\mathbf{2^n}$ possible truth assignments makes the formula true. If yes, output **1**, otherwise output **0**. This brute force procedure takes **exponential(n)** time in general. But is the "SAT problem" tractable, meaning decidable efficiently, i.e., in **polynomial(n)** time? This is equivalent to the well-known **P = NP?** problem of (Cook, 1971), (Karp, 1972), (Levin, 1973).

While the design of novel and efficient algorithms is a key focus of TCS, another unanticipated direction comes of the ability to exploit the power of hard problems. Such problems that cannot be solved efficiently, have been a key insight of TCS. An example is the definition of pseudo-randomness (Yao, 1982). This ability to exploit the power of hardness is novel for mathematics.

## 6.2 The Probabilistic Competition for Conscious Attention and the Influence of the Disposition (a Parameter)

We have made the CTM Up-Tree competition *probabilistic*. We tried to make it deterministic but found that any deterministic competition had to be made increasingly complex to realize important essential properties. For example, if two chunks A and B have nearly equal values $2^{10}$ and $2^{10}+1$, and all other chunks have negligible value 1, and if all values remain unchanged for $\boldsymbol{\tau}$ ticks, then throughout time $\boldsymbol{\tau}$, a deterministic CTM would pay conscious attention to A and not at all to B. Assuming this is unacceptable, deterministic workarounds were found to be frightfully complex.

In the probabilistic ***winner-take-all*** competition described below, on the other hand, a chunk wins with probability proportional to its **f-value**. In the above example, where chunks A and B have (nearly) equal |value|, A and B will each have probability ≈ ½ to win the competition throughout time $\boldsymbol{\tau}$. This means CTM will pay conscious attention to both A and B for approximately $\boldsymbol{\tau}/2$ ticks each. Furthermore, this result will be independent of the location of processors, in particular the winning chunk will be independent of the arrangement of processors on the leaves of the Up-Tree.

This property of a probabilistic winner-take-all competition is not achieved (and possibly not achievable) in tennis and chess tournaments.

We now describe the probabilistic *winner-take-all* competition. First recall that a *chunk* is a tuple,

**<address, time, gist, value; auxiliary information>,**

consisting of the **address** of the originating processor, the **time** the *chunk* was put into the competition, a succinct Brainish **gist** of information, a **valenced value (**to indicate a combination of the importance and with what confidence the originating processor assigns its gist and whether processors should look for ways to increase or decrease the **|value|)**, and some auxiliary information. For the probabilistic CTM, there is good reason to choose the *auxiliary information* to be a pair of numbers that we call **<intensity, mood>**.

*At the start of the competition* (**intensity = |value|** and **mood = value**), this auxiliary information is local information, but as a chunk moves up the Up-Tree it becomes increasingly global. In the winning chunk, the auxiliary information provides "global context".

*To start a competition*, each LTM processor puts a chunk into its leaf node of the Up-Tree. In the probabilistic competition, each non-leaf node of the Up-Tree contains a **coin-toss neuron.** The coin-toss neuron probabilistically chooses the *local winner* of the two competing chunks supplied by the node's two children with probability proportional to their **f-values.** Here **f** is a function mapping each chunk to a single non-negative real number. A *simple*, but natural **f-value** maps each chunk to its **intensity**.

If $C_1$ and $C_2$ are the two competing chunks in a match of the competition, then the coin-toss neuron chooses $C_i$ as local winner of the match with probability $f(C_i)/(f(C_1) + f(C_2))$ (or probability **1/2** if $f(C_1) + f(C_2) = 0$.)

The local winner will move up one level in a single clock tick. The first four parameters of this new chunk are the same as the local winner's. The winner's intensity and mood, however, will be the sum of the winner's and loser's intensity and mood. This is the ***winner-take-all*** policy.

**Theorem.** In a ***winner-take-all* tournament,** one in which every match has a ***winner-take-all* policy,** the probability that a chunk wins the tournament is proportional to its **f-value**, so permuting processor locations will have no effect on which processor wins.

As a chunk moves up the tree, its intensity never decreases while its mood can go up and down. The winning chunk's auxiliary information at the end of the competition will be the pair,

**<sum of all *submitted* chunks' intensities (|values|), sum of all *submitted* moods (values)>.**

Hence, although at the start of the competition each processor has little if any knowledge about the other $\mathbf{N} = \mathbf{2^{24} - 1}$ chunks that are being submitted, the reception of the broadcasted winner provides such information about all processors. In the competition, in addition to providing the winning gist, its given value, and its processor's address, each processor also gains knowledge from the broadcast of the average of all submitted chunks' intensities and moods (on dividing the broadcasted intensity and mood by **N**).[71]

More generally, consider a CTM whose competition employs the following **f-value**:

$$\mathbf{f(chunk) = intensity + d \bullet (mood) \quad with \quad -1 \leq d \leq +1.}$$

Here **d** is a *fixed constant* called CTM's **disposition.**

CTM's *disposition* plays an important part in the competition that selects which chunk will be globally broadcast, and its effect on CTM's behavior.

If the disposition is **d = 0** we say CTM is "level-headed". In this case, **f** is the *simple* intensity discussed earlier. The probability of a submitted chunk "winning" the competition will depend only on its |value|, independent of valence.

If the disposition is **d > 0**, CTM will be "upbeat" in the sense that positively valenced chunks will have a higher probability of winning than negatively valenced chunks of the same |value|. If its

---

[71] **Question.** Why does CTM need a competition? Why not have each processor compute the probability of its chunk winning and then choose the winning chunk based on these probabilities? **Answer.** At the start of the competition each processor has little idea about the other $\mathbf{N} = \mathbf{2^{24} - 1}$ chunks that are being submitted. Without this knowledge, the simple competition is an efficient way to ensure that each chunk gets into STM with probability proportional to its **f-value.**

disposition is $\mathbf{d = +1}$, CTM is manic: only positively valenced chunks can win the competition.[72] The CTM is only conscious of what is positive in its life (as long as anything is positive).

If the disposition is $\mathbf{d < 0}$, CTM will be "downbeat". In the extreme case that $\mathbf{d = -1}$, CTM will be "hopelessly depressed": only negatively valenced chunks can win the competition.[73] In the CTM, there is no way out of this horrible state except with a reboot, i.e., a "shock" to the system that gives it a less extreme disposition.

**Prediction.** In humans, shock therapy will prove as useful for mania as for clinical depression.[74]

## 6.3 Addressing Kevin Mitchell's questions from the perspective of CTM

Here we address Kevin Mitchell's fifteen questions (Mitchell, 2023)[75] from the perspective of the CTM, a robot with a CTM brain.

Our answers refer to and supplement our Overview (Section 2). They deal *only* with the CTM model, meaning an entity with a CTM brain. That entity could be a human if it has a CTM brain. These answers say nothing about other models. They say nothing about whether or not a worm is conscious unless the worm has a CTM brain. From here on, unless we say otherwise, everything we have to say is about the CTM model. We do not claim our answers to be complete or to be the only ones possible.

---

[72] – Unless of course no chunks have positive valence, in which case a negatively valenced chunk must get into STM, and each such chunk will have probability $\mathbf{2^{-24}}$, independently of its value, to get to STM.

[73] – Unless of course no chunks have negative valence, in which case a positively valenced chunk must get into STM, and each such chunk will have a probability of $\mathbf{2^{-24}}$, independently of its value, to get to STM.

[74] In humans, electroconvulsive therapy (ECT) is used primarily for extreme depression (**d=-1** in CTM), not for mania. We suggest there is a fear that since shock therapy raises the mood of depressives, it might raise the mood of maniacs. The CTM theory suggests the contrary, that it should be just as useful for humans in the manic state (**d=+1** in CTM). See (Elias, Thomas, & Sackeim, 2021).

[75] Many of Mitchell's questions are in fact a collection of intertwined questions.

In this paper we defined two related notions of consciousness in CTM, *conscious attention* (Formal Definition 1 in section 2.2) and *conscious awareness* (Formal Definition 9 in section 2.3.2.2). Before answering the questions, we review these formal CTM definitions and related axioms:

> ***Conscious attention*** in **CTM** is the simultaneous *reception* by all LTM processors of the global *broadcast* of the winner of the competition for **STM** that started at time **t**.
>
> ***Conscious awareness*** of that winner, **<p, t, g, v; aux>**[76]**,** occurs when **CTM** *unpacks* this chunk, i.e., all **LTM** processors simultaneously (in the same tick) *unpack* the *meanings* of all Brainish words[77] appearing in gist **g** and take note of its *value* **v**.

The *global reception* by all LTM processors of a "sensory chunk" **<p, t, g, v; aux>** evokes a *unitary experience* in CTM (the Global Broadcast Axiom 1, in section 2.2) and the Brainish word **<p, t>** for this "experience" is created. If in addition CTM *unpacks* this chunk, a unique *subjective/sensory experience* is *evoked* (the Unpacking Axiom 2, in section 2.3).

In the following, Mitchell's fifteen questions, KM1, ..., KM15 are printed in **bold.** Our answers follow in non-bold print.

**KM1. What kinds of things are sentient? What is the basis of subjective experience and what kinds of things have it? What kinds of things is it like something to be?**

Again, our answers deal *only* with the CTM model.

**What kinds of things are sentient?** We take "sentience" to mean the ability to "sense" and "feel" things. In this sense, CTM is sentient: Sensory processors in CTM convert sensory inputs into chunks. If CTM pays *conscious attention* to a "sensory chunk" **<p, t, g, v; aux>**, a *unitary experience* is *evoked* (consequence of Axiom 1) and the Brainish word **<p, t>** for this "experience" is created. If in addition CTM becomes *consciously aware* of this chunk, a unique *subjective/ sensory experience* is *evoked* (by Axiom 2). Later, when CTM becomes *consciously aware* of a chunk containing the word **<p, t>**, a modification of the original *sensory experience* will be *evoked*.

---

[76] Recall, a chunk is given by a tuple, **<p, t, gist, v; aux>,** where **p** is the address of the processor that created the chunk, **t** is the time the chunk was put into competition for **STM**, **gist** is the succinct information in Brainish the processor is submitting, **v** is the valenced number the processor has given the gist, and **aux** is auxiliary information (not necessarily given by processor **p**).

[77] *Brainish words* are pointers **<p, t>** to broadcasted chunks, **<p, t, g, v; aux>**, their *meanings*. Broadcasted chunks are stored by all LTM processors, thus creating a *global dictionary* in each LTM. For a processor to *unpack* a word **<p, t>** means for it to look into its dictionary to *retrieve* and *read* the chunk **<p, t, g, v; aux>**.

**What is the basis of subjective experience and what kinds of things have it?** The ability of CTM to construct models of its worlds (inner and outer) is crucial for its *subjective experiences*. Brainish *labeled sketches* (Definition 5 in section 2.3.2.1) are *created* in these models and *evolve* throughout the life of CTM. The labels succinctly indicate what CTM learns, senses or feels about the sketches' referents. In the case of a red rose, the Brainish labeled sketch might be an image, fused with its odors, smooth touch, and so on. When CTM becomes *consciously aware* of these labeled sketches, subjective experiences are evoked.

**What kinds of things is it like something to be?** Formally, the Model-of-the-World processor (MotWp) and the Model-of-the-World (MotW) it creates with Brainish labeled sketches play essential roles in "what it is like" to be a conscious CTM (section 2.3). The label SELF applied to a sketch in the MotW indicates that that particular sketch's referent is felt as a part of, or the whole of, CTM itself. The label PAIN indicates that that particular referent is in pain. Labels like PAIN (sections 2.3.1.1 and 2.3.2.3) and SELF (sections 2.3.1.2 and 2.3.1.3) are Brainish words **<p, t>,** which are pointers to chunks **<p, t, gist, v; aux>** in which experiences (like PAIN and SELF) arise in "meaningful" ways (section 2.3.2.2). *When CTM becomes consciously aware of these labeled sketches*, for example of the labeled sketch <KNEE><SELF◦PAIN>, *CTM will experience* itself in pain.

**KM2. Does being sentient necessarily involve conscious awareness? Does awareness (of anything) necessarily entail self-awareness? What is required for 'the lights to be on'?**

**Does being sentient necessarily involve conscious awareness?** "Sentience" is the ability to "sense" and "feel" things. So yes, in CTM sentience *is* conscious awareness.

**Does awareness (of anything) necessarily entail self-awareness?** No, not unless the current MotW has a sketch of the thing labeled SELF and CTM becomes *consciously aware* of that labeled sketch. For example, the sketch of the newborn CTM in the MotW is just an empty blob at first, representing the "world". In time, that world model will include a rough sketch of CTM labeled $SELF_0$. Conscious awareness of that labeled sketch marks the beginning of self-awareness, which eventually develops into full-blown self-awareness.

**What is required for 'the lights to be on'?** In CTM, the *lights come on* gradually, in lockstep with the MotW getting populated with Brainish labeled sketches. (For more on this, see our answers to KM3 and KM4.)

**KM3. What distinguishes conscious from unconscious entities? (That is, why do some entities have the *capacity* for consciousness while other kinds of things do not?) Are there entities with different degrees or kinds of consciousness or a sharp boundary?**

Again, we replace "entities" with "CTMs".

**What distinguishes conscious from unconscious entities? (That is, why do some entities have the *capacity* for consciousness while other kinds of things do not?)** All CTMs have the *capacity* to be conscious. Absent a broadcast system there is no *conscious attention*. Absent the ability to unpack a broadcasted chunk there is no *conscious awareness* (subjective consciousness).

**Are there entities with different degrees or kinds of consciousness or a sharp boundary?** In CTM, we have distinguished between *conscious attention* and *conscious awareness*. Alison Gopnik (Gopnik, 2007) has said that babies are more conscious than adults. With respect to *conscious attention*, this is true for infant CTMs: before links have formed, all communication goes through STM and is broadcasted to all processors. This is conscious attention. Conscious awareness, however, comes more slowly, with *the co-evolution of Brainish and world models* (section 2.3.1).

CTM can exhibit different *degrees of consciousness*:

1. CTM can be consciously aware when awake or dreaming. It is not conscious when in *deep sleep*. (See our answers to KM4 for discussions of Sleep and Dream processors.)

2. Faulty processors or faults in the Up-Tree competition can affect what gets into STM, hence affect both conscious attention and conscious awareness. For example, a faulty CTM can have blindsight, meaning it can do some things that are normally done with conscious sight, but without any *conscious sense* that it can see (Blum & Blum, 2022). This will happen if the Vision processor fails to get its chunks into STM (e.g., relevant branches in the Up-Tree are broken or the Vision processor fails to give high enough |value| to its chunks), while previously formed links enable visual information to be communicated to the Motor processors.[78]

3. We have defined *conscious awareness* in terms of the first level unpacking (section 2.3.2.2). Deeper unpackings may evoke higher degrees of conscious awareness. Additionally, *conscious awareness* of being consciously aware is a *higher order* consciousness.

**KM4. For things that have the capacity for consciousness, what distinguishes the *state* of consciousness from being unconscious? Is there a simple on/off switch? How is this related to arousal, attention, awareness of one's surroundings (or general responsiveness)?**

To answer, we focus mostly on CTM's sleep, dream, and awake states**.**

**For things that have the capacity for consciousness, what distinguishes the *state* of consciousness from being unconscious?**

First, what is a state? For each time **t**, it is a complete description of the CTM at time **t**. A *conscious state* in CTM is an *awake* or *dream* state. A *dreamless sleep* state is an *unconscious state*. Assuming a functioning CTM, the content of a global broadcast may determine whether an upcoming state will be conscious or unconscious.

For example, an *unconscious state* will occur when a *Sleep processor* generates a sleep chunk with a "sufficiently high" |value|, one with high enough value to get into STM and keep other chunks out. This will prevent CTM from interacting with the world.[79] This is the *unconscious deep sleep* state.[80] In this state, CTM is not aware of its surroundings, though it might be *aroused* by pangs of intense hunger, other pains, a very loud explosion, and so on. This occurs, for example, when these pangs, pains, and sounds overwhelm the Sleep processor with an even larger |value|. CTM can get out of the unconscious deep sleep state if the Sleep processor lowers its submitted chunks' |value|s enough to get into (for example) the dream state.

---

[78] In the human visual cortex, the dorsal stream of vision is unconscious; the ventral stream is conscious. Studies on blindsight suggest that communications via the unconscious dorsal stream account for the surprising blindsight ability of visually impaired people (Tamietto & Morrone, 2016).

[79] Something like this can also happen in total depression and catatonia in CTM, and slow-wave (non-REM) sleep in humans.

[80] Even in deep sleep, however, a CTM can still carry out tasks (utilizing unconscious communication between processors via links) but without conscious attention and therefore without conscious awareness.

Another seemingly unconscious state occurs when all chunks submitted to the competition for STM have the same |value|. In CTM's *winner-take-all* probabilistic competition, chunks reach STM with probability proportional to (a *monotonically increasing function* of) the chunk's |value|. If all chunks have the same |value|, they all have equal probability $1/2^{24}$ to get to STM. In that case, for as long as all values stay the same (don't change), the chunks flit in and out of STM *at random.* The CTM loses anything resembling a sustained attention (like Robbie the Robot in *Forbidden Planet,* (Nayfack, 1956)). To calm things down, a Gauge processor (e.g., the Sleep processor) may submit a high enough valued chunk to overrule the others.

**Is there a simple on/off switch?** The Sleep processor is an example of an *on/off switch* for consciousness in CTM. (The Sleep processor, like Gauge or pain processors, is not generally under conscious control.)

**How is this related to arousal, attention, awareness of one's surroundings (or general responsiveness)?** Returning to the unconscious sleep state, when the |value| of the Sleep processor's chunk drops sufficiently - but not enough to let input-output chunks enter STM - CTM's *Dream processor* can take over, enabling chunks of a dream to emerge and enter STM. We consider the Dream state to be a *conscious state*. When the |value| of the Sleep processor's chunks drops further, CTM wakes up.

---

Similarly, if CTM's broadcast station is turned off, there is no conscious attention and no conscious awareness. If this occurs after links and unconscious algorithms have been created, then while the broadcast station is down, CTM is an *unconscious zombie*. It can still function unconsciously with algorithms and links it acquired when it was conscious.

**KM5. What determines what we are conscious of at any moment? Why do some neural or cognitive operations go on consciously and others subconsciously? Why/how are some kinds of information permitted access to our conscious awareness while most are excluded?**

**What determines what we are conscious of at any moment?** In CTM, the competition for STM determines which one of ~ten million chunks that were entered into the competition CTM will become conscious. Conscious attention enables all processors to focus on the same thought. When a task needs to be accomplished, all processors attempt it for an amount of time determined by the value of the task, the processor's free time (section 2.1.7 (5)), and the processor's knowledge, ability, and motivation.

**Why do some neural or cognitive operations go on consciously and others subconsciously?** [**and others unconsciously]?**

*Conscious activity* in CTM is deliberate. It is triggered by information that goes through STM and is broadcast to all LTM processors.

We call *subconscious* whatever is submitted to the (Up-Tree) competition for STM that does not win. Such chunks are precursors for consciousness, in Freud's "anteroom" (Freud, 1915 (translated 1920, G. Stanley Hall)).

We call *unconscious* those operations that do not involve STM:

1. Operations within each LTM processor. These operations enable processors to do what they need to do efficiently with minimal distractions, including creating chunks.
2. Communication between LTM processors via links.

*Unconscious activity* is quicker than *conscious activity*. *Unconscious activity* can become *subconscious*, and *subconscious activity* can become *conscious*.

**Why/how are some kinds of information permitted access to our conscious awareness while most are excluded?** CTM pays conscious attention to all broadcasted chunks. In addition, it becomes consciously aware of those particular (broadcasted) chunks that have sufficiently high |value|. Those are the chunks that for whatever reason are likely to contain questions from, or information for, the LTM (unconscious) processors.

**KM6. What distinguishes things that we are currently consciously aware of from things that we *could be* consciously aware of if we turned our attention to them, from things that we *could not be* consciously aware of (that nevertheless play crucial roles in our cognition)?**

Conscious awareness in CTM has to do with *subjective experience*, which is *evoked* when a broadcasted chunk is unpacked (See section 2.3.2.2 and KM2.), most likely because it has a high |value|.

What things, though important for cognition, *cannot* enter consciousness? Here are several examples from CTM:

1. Things that must be done so quickly that the communication necessary to do the thing hasn't the time to go through STM. For example, CTM must quickly swerve away from an oncoming car while riding its bike.
2. Things like **abc** whose doing would take away from a more important thing called **xyz**, when there is time to consciously attend to only one of **abc** and **xyz.**
3. Specific details of the predictive dynamics (prediction, feedback, learning) built into every processor. Specific constants that are optimized (unconsciously) by those algorithms.

**KM7. Which systems are required to support conscious perception? Where is the relevant information represented? Is it all pushed into a common space or does a central system just point to more distributed representations where the details are held?**

Again, our answers presume a system built and functioning like CTM.

**Which systems are required to support conscious perception?** Assuming conscious *perception* is *conscious awareness*, the required major systems are the Up-Tree competition to select a winning chunk, the single broadcasting station for broadcasting the winning chunk to all processors, the systems in every processor for *unpacking* that broadcasted chunk (including algorithms to access and examine the meaning of Brainish words). Unpacking generates the CTM's conscious perception.

**Where is the relevant information represented? Is it all pushed into a common space or does a central system just point to more distributed representations where the details are held?**

Information in CTM is represented by Brainish words **<p, t>** and the broadcasted chunks **<p, t, gist, value; aux>** they point to, which give those words their meaning. These chunks (and implicitly the words that point to them) are entered into identical global dictionaries that are created and maintained by every processor and stored in each processor's personal "global" memory.

The Model-of-the-World processor is distributed across all LTM processors and creates labeled sketches of referents to CTM's inner and outer worlds. When a labeled sketch (as the gist in a chunk) is broadcast, it (like all chunks) is stored in each and every processor's global dictionary.

**KM8. Why does consciousness feel unitary? How are our various informational streams bound together? Why do things feel like *our* experiences or *our* thoughts?**

**Why does consciousness feel unitary? How are our various informational streams bound together?** *Conscious attention* in CTM is the result of all processors simultaneously receiving the broadcasted information. This evokes in CTM a *unitary* experience (Axiom 1). Additionally, for example, when CTM becomes *consciously aware* of the sketch of a rose in its MotW with its *fused multimodal Brainish* label RED_COLOR ∘ SWEET_SMELL, it (the CTM) "sees" and "smells" a sweet red rose (Axiom 2). This broadcasted information feels unified because it is fused in the broadcasted gist, and it is received and unpacked simultaneously by all processors.

**Why do things feel like *our* experiences or *our* thoughts?** When CTM becomes *consciously aware* of a thought (i.e., a sketch of a thought in the MotW) labeled SELF, CTM will *experience* that thought as its own (section 2.3 and Axiom 2).

**KM9. Where does our sense of selfhood[81] come from? How is our conscious self related to other aspects of selfhood? How is this *sense of self* related to actually *being a self*?**

Here again, world models, with their learned-from-experience Brainish labeled sketches, determine CTM's sense of self and other aspects of selfhood. Over time, the MotW's sketches of CTM become labeled with a variety of Brainish words. For this question, the labels SELF and FEEL are particularly important. When CTM becomes *consciously aware* of such a Brainish labeled

[81] Selfhood "is the experience of being a distinct, holistic entity capable of global self-control and attention, possessing a body and a location in space and time." (Blanke & Metzinger, 2009)

sketch it will feel that that sketch represents a part or the whole of itself. Known pathologies of selfhood occur when any one or both of these labels are missing, or when sketches are mislabeled.[82] CTM's interaction with its inner and outer worlds through sensors and actuators connects its *sense of self* with actually *being a self.*

**KM10. Why do some kinds of neural activity feel like something? Why do different kinds of signals feel different from each other? Why do they feel specifically like what they feel like?**

**Why do some kinds of neural activity feel like something?** "Neural activity" in CTM is activity in the sensors, the actuators, the LTM processors, the links between them, the Up-Tree and the broadcasting system. Each sensory processor gets input from specific sensors, turns that input into a Brainish gist (a Brainish representation of that input), puts that gist in a chunk, and puts that chunk in the competition for STM. If and when a broadcast of that chunk is *received* and *unpacked* by all processors, it *evokes* in CTM a feeling that depends on the originating sensory processors' chunk (Axiom 2).

**Why do different kinds of signals feel different from each other?** In CTM, each and every *broadcasted* chunk **<p, t, ... >** defines a new word **<p, t>** and, if simultaneously *unpacked*, an associated new experience. Since any two chunks differ in at least their **<p, t>**, and thus information about the processors that created the chunks, the *evoked* feelings are different.[83]

**Why do they feel specifically like what they feel like?** In the MotW, sketches of a red rose and a red fire engine will both get the Brainish label RED. Each of these sketches will also get labels that the other does not get. The feelings are determined by the set of labels, and different sets of labels *evoke different subjective experiences* (distinct *platonic qualia*) when CTM becomes consciously aware of them. For another, the *labeled sketches*, <ROSE><RED> and <FIRE-ENGINE><RED> are distinct and so *will evoke distinct subjective experiences* (distinct *enacted qualia*).

---

[82] Some human-like examples of pathologies due to mislabeling include body integrity dysphoria and asomatognosia (when SELF is missing from some body part), phantom limb syndrome (when an amputated arm is still labeled SELF), Cotard's syndrome (when SELF and FEEL are missing from the representation of oneself in the MotW)**,** paranoia (when a friend is labeled SPY), ….

[83] A similarity measure to indicate how close or different two feelings are in CTM has yet to be defined.

**KM11. How do we become conscious of our own internal states? How much of our subjective experience arises from homeostatic control signals that necessarily have valence? If such signals entail feelings, how do we know what those feelings are about?**

**How do we become conscious of our own internal states?** CTM becomes conscious of its own (internal) state when a global broadcast of a chunk with a gist of that state is received. For example, in section 2.3.1.2, we indicated how the newborn CTM would know it needs something (e.g., fuel) when a high |value| negatively valenced chunk from a processor (e.g., its Fuel Gauge processor, a homeostat) reaches STM and is broadcast from it.

**How much of our subjective experience arises from homeostatic control signals that necessarily have valence? If such signals entail feelings, how do we know what those feelings are about?** In response to the reception of the negatively valenced LOW_FUEL chunk submitted by the Fuel Gauge processor (a homeostat), the sketch of CTM in the MotW will be labeled <HUNGRY>. The broadcast and unpacking of a negatively valenced chunk with the Brainish labeled sketch <CTM><HUNGRY> will evoke in CTM the subjective experience of "hunger". An actuator will eventually connect CTM's fuel intake to a Fuel Source. Assuming the fuel transfer is successful, the Fuel Source sketch in the MotW will eventually become labeled <RELIEVES_HUNGER ○ PROVIDES PLEASURE >. The broadcast and unpacking of a positively valenced chunk with this Brainish labeled sketch will create the fused subjective experience of "hunger relieved ○ pleasure". CTM will learn that "the Fuel Source is a hunger reliever and pleasure provider" and that "CTM will experience pleasure if it gets food when hungry."

The above is an example of homeostasis in CTM. In particular, it is an example of how CTM's subjective experiences arise from homeostatic control signals that have valence and how CTM "knows" what those feelings *are about*. See also our answer to the next question.

**KM12. How does the aboutness of conscious states (or subconscious states) arise? How does the system know what such states refer to when the states are all the system has access to?**

**How does the about-ness of conscious states (or subconscious states) arise?** We define the ***conscious state*** at time **t** to be the chunk received by all processors at that time. So conscious states are chunks that are received simultaneously by all processors.

Suppose **chunk1** (received by all processors at time **t1**) contains a gist describing the world that CTM faces from the top of a down-staircase.

Suppose **chunk2** (received at time **t2**) predicts the effect of starting down the stairs. This prediction is done by the MotWp simulating that action[84] (in the MotW). **Chunk3** (received at time **t3**) then sends a command to CTM's walk actuator to go down the stairs. **Chunk4** (received at time **t4**) reports back.

That original starting state (chunk1), the prediction state (chunk2), the action state (chunk3), and the report state (chunk4) provide the desired (conscious) *about-ness* for the experience.

While CTM is not conscious of unconscious or subconscious states, they nevertheless can influence conscious states. As for subconscious states, they could have been conscious but for the luck of the draw; their variants could still become conscious. As for unconscious states, they determine what chunks get submitted to the competition.

**How does the system know what such states refer to when the states are all that the system has access to?** The CTM only has conscious access to *conscious states*, broadcasted chunks received by all processors. Such a chunk's gist is a Brainish description of what CTM knows of its world. If CTM *unpacks* this chunk, i.e., becomes *consciously aware* of this state, CTM will know what the state refers to.

**KM13. What is the point of conscious subjective experience? Or of a high level common space for conscious deliberation? Or of reflective capacities for metacognition? What adaptive value do these capacities have?**

**What is the point of conscious subjective experience?** Without it, CTM is not compelled to act appropriately. A CTM that has pain knowledge, *knows* that pain can damage or destroy it, but does not subjectively experience it, is not motivated to respond normally to the pain, to take care of itself. Such a person, robot or CTM that lacks the ability to *subjectively feel* the agony of pain has *pain asymbolia.*

[84] This is similar to the kind of simulation that the MotW does in a dream sequence.

**What is the point of a high level common space for conscious deliberation?** We view the CTM's global theater as the "high level common space for conscious deliberation". We view the STM buffer and broadcast station as its focal point: only one processor can write in it at any time (i.e., at each clock tick, only one chunk can win the competition); every processor has a chance to write in it. Global broadcasting focuses all processors on the same thought (chunk). With the reception of a broadcast, all processors have an opportunity to contribute to its understanding and, if it's a problem, its solution.

For example, suppose the broadcast is a high |value| negatively valenced chunk, a kind of "scream", indicating the problem: "Hungry! Need to fill the fuel tank." To deal with this situation, different processors might suggest different responses, like vocal screams, vocal cries, wake up calls, other actions.

**What is the point of reflective capacities for metacognition?** Reflective capacities enable CTM to treat itself with all the planning tools it uses to treat other referents.

**What adaptive value do these capacities have?** They enable CTM to adapt. For example, how does CTM adapt to hot weather? Suppose the broadcast is: "It's too hot: find a way to cool down!" One processor suggests: "Find a cool place to go to." Another suggests: "Get some water to drink." Different processors have different ideas what to do. The competition will select one of the suggestions for broadcast. An actuator can then carry out the suggestion. If the suggestion doesn't resolve the situation as implicitly predicted, another suggestion will be requested. And so on.

**KM14. How does mentality arise at all? When do information processing and computation or just the flow of states through a dynamical system become elements of cognition and why are only some elements of cognition part of conscious experience?**

**How does mentality arise at all?** We restate this as: **How does the capacity for intelligent thought arise?** While all ~ten million LTM processors are independent and unlinked at $\mathbf{t = 0,}$ the Up-Tree competition's chunk selection and its broadcast enable and encourage processors with different abilities and ideas, related or not, to work together to solve problems. The whole can be more than the sum of its parts. This makes for creative intelligent thinking and problem solving. See section 2.4, where we discuss CTM as a framework for Artificial General Intelligence (AGI).

**When do information processing and computation, or just the flow of states through a dynamical system, become elements of cognition?** Viewing CTM as a dynamical system, the process of competition, selection of a winner, and reception of the winning chunk by all LTM processors (resulting in conscious attention to it) is an element of cognition. Intrinsic to this process are the cycles of prediction, feedback and learning. If all processors *unpack* the received chunk, then CTM becomes consciously aware of the information, a deeper element of cognition.

**Why are only some elements of cognition part of conscious experience?** Some elements of cognition can be performed by LTM processors working alone or with others using just their previously learned knowledge and algorithms, without going through STM. Those elements of cognition are unconscious. Processors that *do* need to search for information – or for a computational capability - can broadcast their need.[85] That broadcast to all processors is an essential ingredient of conscious cognition.

**KM15. How does conscious activity influence behavior? Does a capacity for conscious cognitive control equal "free will"? How is mental causation even supposed to work? How can the meaning of mental states constrain the activities of neural circuits?**

**How does conscious activity influence behavior?** As an example, by bringing together seemingly unrelated ideas, conscious activity can generate wildly creative approaches for solving problems. As another example, conscious attention to a chunk of sufficiently high |value| and negative valence interrupts the work of all processors and forces them to pay their maximum (*free*) time addressing the chunk's gist. This happens, for example, in a fight, flight, or freeze confrontation, when the best choice must be made in a very limited time.

**Does a capacity for conscious cognitive control equal "free will"?** CTM's ability to assess a situation, meaning to consider various possible actions and predict the consequences of each, given the resources it has, gives CTM an excellent reason to believe it has "free will". To see this, imagine CTM playing a game of chess. When and while CTM has to decide which of several possible moves

---

[85] Like the processor that asks, "What her name?" (See footnote in section 2.1.2).

to make, it knows it can choose - is free to choose - whichever move has the greatest utility for it. That is “free will”. (Blum & Blum, 2022)

**How is mental causation even supposed to work?** Mental causation is what we have called the *power of thought* (section 2.3.1.3). In CTM, the MotWp and the MotW are fundamental to *mental causation*. To decide whether to act in the world, an LTM processor (part of the MotWp) simulates that action in the MotW and from that decides whether or not to go ahead with it. If it does, then CTM looks to see if the requested action got accomplished as predicted. (In any case, whether the prediction was followed closely or not, the LTM processor corrects/improves the prediction algorithm, unconsciously, using reinforcement learning).

For example, suppose the infant CTM discovers that it can somehow move its left leg. It becomes aware through its sensors that “willing” the movement of that leg (with the power of thought) is successful. For comparison, it will discover via its sensors that it cannot pick up a rock, Yoda style, with the power of thought. Moving the leg or lifting the rock are “willed” in the world by performing the action in and through the MotW. Sensors must verify if the act has been successful. If it has, that is mental causation.

**How can the meaning of mental states constrain the activities of neural circuits?** As an example, previously noted, sufficiently high |value| negatively valenced chunks (a mental state) cause all processors to put their current activity on hold and (to the extent possible) work to lower the |value|.

*****

We’ve now come to the end of Kevin Mitchell’s questions. Mitchell ends his blog with the words, “If we had a theory that could accommodate all those elements and provide some coherent *framework*[86] in which they could be related to each other – not for providing all the answers but just for asking sensible questions – well, that would be a theory of consciousness.” (Mitchell, 2023) This is our goal for the theory of the Conscious Turing Machine.

[86] Italics ours.